\documentclass{article}

\usepackage[final]{neurips_2023}
\usepackage[utf8]{inputenc}
\usepackage[T1]{fontenc}
\usepackage{lmodern}
\usepackage{xcolor}

\usepackage{graphicx}
\usepackage{booktabs}
\usepackage{multirow}
\usepackage{caption}
\usepackage{subcaption}
\usepackage{float}

\usepackage{amsmath, amssymb, amsfonts}
\usepackage{bm}

\usepackage{hyperref}
\hypersetup{
    colorlinks=true,
    linkcolor=blue,
    citecolor=blue,
    filecolor=magenta,      
    urlcolor=blue,
}

\begin{document}
\begin{center}
    {\LARGE \textbf{ObitoNet: Multimodal High-Resolution Point Cloud Reconstruction}} \\[1em]
    
    \begin{tabular}{ccc}
        \textbf{Vinay Lanka} & \textbf{Apoorv Thapliyal} & \textbf{Swathi Baskaran} \\
        M.Eng in Robotics & M.Eng in Robotics & M.S. in Data Science \\
        University of Maryland & University of Maryland & University of Maryland \\
        College Park & College Park & College Park \\
        \texttt{vlanka@umd.edu} & \texttt{apoorv10@umd.edu} & \texttt{swathib@umd.edu} \\
    \end{tabular}
\end{center}


\begin{abstract}
We present \textbf{ObitoNet}, a multimodal framework for high-resolution point cloud reconstruction that effectively fuses image features and point cloud data through a Cross-Attention mechanism. Our approach leverages \textbf{Vision Transformers (ViT)} to extract rich semantic features from input images, while a \textbf{point cloud tokenizer} —utilizing \textbf{Farthest Point Sampling (FPS)} and \textbf{K-Nearest Neighbors (KNN)}—captures local geometric details. These multimodal features are combined using a learnable \textbf{Cross-Attention module}, which facilitates effective interaction between the two modalities. A \textbf{transformer-based decoder} is then employed to reconstruct high-fidelity point clouds. The model is trained with Chamfer Distance (L1/L2) as the loss function, ensuring precise alignment between reconstructed outputs and ground truth data. Experimental evaluations on standard benchmark datasets, including ShapeNet, demonstrate that \textbf{ObitoNet} achieves comparable performance to state-of-the-art methods in point cloud reconstruction. The results highlight the robustness and efficacy of our multimodal feature fusion approach in generating high-resolution point clouds, even from sparse or noisy inputs. \\
\textbf{Github link:} \href{https://github.com/vinay-lanka/ObitoNet/}{https://github.com/vinay-lanka/ObitoNet/}

\end{abstract}

\section{Introduction}
Point clouds are collections of discrete data points in three-dimensional space, typically represented as \((x, y, z)\) coordinates. They serve as a critical data structure for representing 3D objects and scenes in fields like computer vision and robotics. However, point clouds are inherently sparse, unordered, and irregular compared to structured data formats like images or grids, posing significant challenges for deep learning-based methods. Consequently, reconstructing high-resolution point clouds from sparse or noisy inputs remains a major research challenge.

Recent advancements in computer vision have shown the effectiveness of ViTs in capturing semantic information from images using patch-wise self-attention. For point-cloud data, methods like PointNet and PointNet++ provide effective frameworks for processing unordered 3D points. Techniques such as Farthest Point Sampling (FPS) and K-Nearest Neighbors (KNN) enable the grouping and sampling of points to preserve local geometric structures. While image and point-cloud-based methods perform well independently, their combination has the potential to enhance high-resolution reconstruction. The core idea is to guide the model in bridging gaps within the point cloud by leveraging complementary features extracted from images. This is achieved through Cross-Attention, which seamlessly integrates image and point-cloud modalities, enabling the generation of detailed and high-resolution point clouds. Our approach consists of three major steps:

\textbf{Feature Extraction.} For images, we leverage \textbf{ViTs} to extract rich semantic features by treating images as sequences of patches and applying patch-wise self-attention. For point clouds, we utilize a tokenizer that combines \textbf{FPS} to select representative points with \textbf{KNNs} to group local neighborhoods, effectively capturing fine-grained geometric details while maintaining spatial structures.

\textbf{Multimodal Fusion.} The extracted image and point cloud tokens are aligned and fused using a Cross-Attention mechanism, effectively combining the semantic context of images with the geometric details of point clouds to enrich the quality of the feature representation.

\textbf{High-Resolution Reconstruction.} A transformer-based decoder processes the combined features to produce high-quality point clouds, improving spatial details and accurately reconstructing geometric structures.

\section{Related Work}
Reconstructing high-resolution 3D point clouds is a core challenge in computer vision and 3D processing. Early methods, like \textbf{PointNet} [Qi et al., 2017]\cite{pointnet}, introduced frameworks for unordered point sets but struggled with capturing fine-grained geometric details and handling high-resolution data efficiently. Subsequent approaches, such as \textbf{PointNet++} [Qi et al., 2017b]\cite{pointmae}, improved local feature aggregation, while voxel-based methods [Graham et al., 2018]\cite{conv} enabled efficient processing but faced memory limitations. Building on these foundations, our work tackles the challenge of fine-grained reconstructions by leveraging Cross-Attention to integrate spatial and visual features.

Multimodal approaches, like \textbf{multi-view CNNs} [Su et al., 2015]\cite{multiviewcnn}, demonstrate the value of using visual data to complement sparse 3D inputs. Similarly, the success of \textbf{Transformers} [Vaswani et al., 2017]\cite{attention} in feature fusion has inspired their application in point cloud tasks. For example, \textbf{PointBERT} [Yu et al., 2021]\cite{pointbert} and \textbf{PointMAE} [Pang et al., 2022]\cite{pointmae} showcase the effectiveness of transformer-based pretraining for point cloud representation and reconstruction. 

Building on these advances, our proposed approach, \textbf{ObitoNet}, leverages Cross-Attention modules to integrate high-resolution spatial and visual features for accurate point cloud reconstruction, aligning with recent methods like \textbf{Neural Radiance Fields (NeRF)} [Mildenhall et al., 2020]\cite{nerf}, while maintaining computational efficiency.

\begin{figure}[ht]
    \centering

    \begin{subfigure}[b]{0.8\textwidth} 
        \centering
        \includegraphics[width=\textwidth]{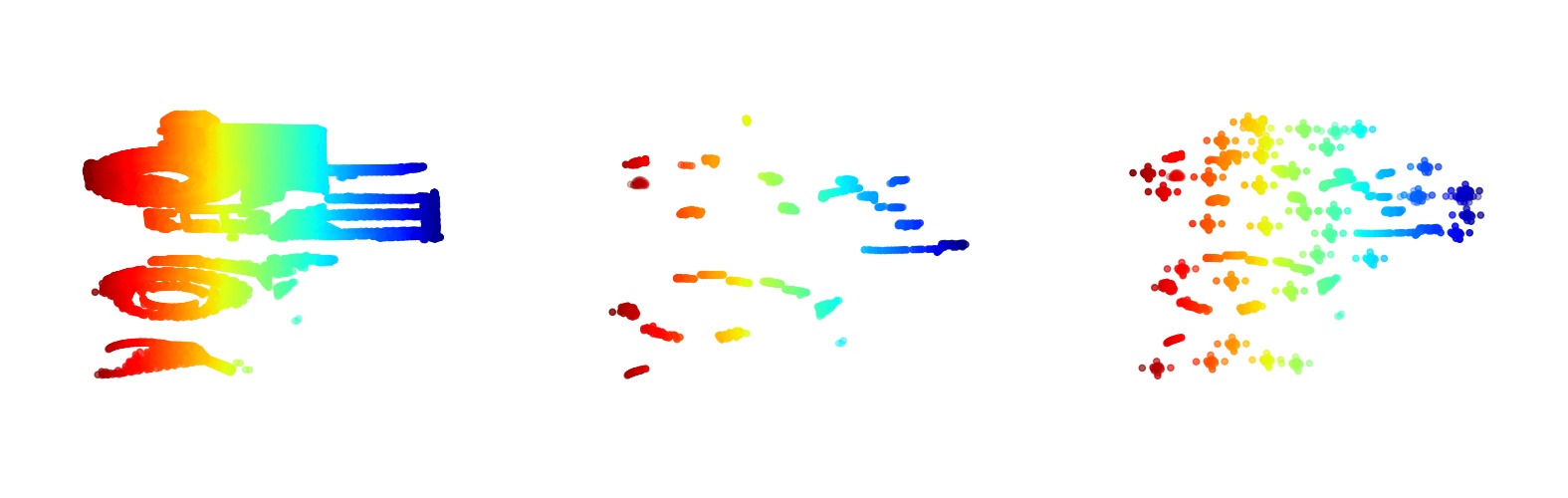}
        \includegraphics[width=\textwidth]{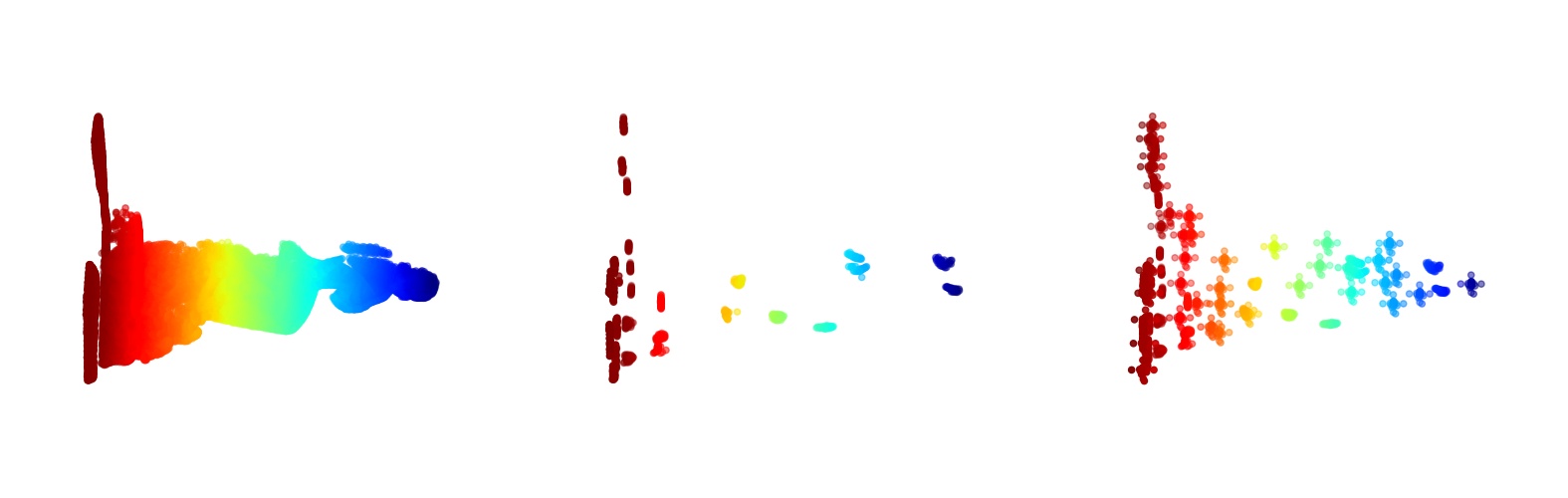}
        \caption{Point Cloud Visualizations}
        \label{fig:pointclouds(1)}
    \end{subfigure}

    \caption{PointMAE pointcloud reconstruction sample outputs}
    \label{fig:larger_pointcloud_visualization}
\end{figure}

\section{Methodology}
This section outlines the architecture, dataset generation, and training strategies employed in the proposed \textbf{ObitoNet} framework. The model integrates multimodal data from images and point clouds using a modular design, consisting of an image tokenizer, a point cloud tokenizer, and a cross-attention-based decoder. The following subsections detail the model's components, the dataset preparation process, and the training methodology, emphasizing how each step contributes to leveraging the complementary strengths of visual and geometric data. We also describe the loss function, training order, and innovations that ensure robust and accurate reconstruction. The modular nature of the framework enables both end-to-end learning and component-wise optimization, making \textbf{ObitoNet} versatile for a range of applications in 3D processing and computer vision.

\subsection{Architecture} 
\begin{figure}[ht] 
    \centering 
    \includegraphics[width=1.0\textwidth]{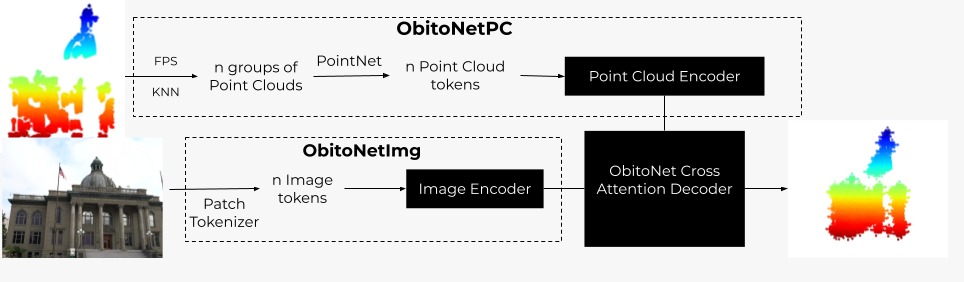} 
    \captionsetup{justification=centering}
    \caption{Overview of the proposed pipeline. ObitoNet integrates image features with point cloud tokens for robust reconstruction.} 
    \label{fig:pipeline} 
\end{figure}

As shown in Figure~\ref{fig:pipeline}, \textbf{ObitoNet} consists of three key components: an image token extractor, a point cloud token extractor, and a Cross-Attention-based decoder that reconstructs the point cloud.

\begin{enumerate} 
\item \textbf{ObitoNetImg:} Employs a standard \textbf{Vision Transformer (ViT)} model to extract rich and meaningful image tokens, which serve as semantic representations for downstream processing.
\item \textbf{ObitoNetPC:} The point cloud data is segmented and grouped into smaller clusters using FPS to select representative points and KNN to capture local neighborhoods, enabling effective processing of fine-grained geometric structures. 
\item \textbf{ObitoNetCA:} A \textbf{Cross-Attention} mechanism integrates image tokens with point cloud tokens, leveraging visual cues to enhance geometric reconstruction and refine spatial details.

\end{enumerate}

\vspace{-12pt}

\subsection{Dataset Generation} 
For evaluating the performance of \textbf{ObitoNet}, we used the \href{https://www.tanksandtemples.org/}{\textit{Tanks and Temples}} dataset, a benchmark for multi-view 3D reconstruction tasks. This dataset provides high-quality 3D point clouds and corresponding 2D RGB images of real-world scenes, offering rich geometric and visual data for training and validation.

To create paired images and point clouds, we projected the 3D point cloud data onto a 2D plane, generating a reference image representation. This reference was compared against existing images in the \textit{Tanks and Temples} dataset using similarity metrics. After testing several approaches, we found that \textbf{CLIP}, a vision-language model, was the most effective for identifying matches based on visual and contextual features. Images and point clouds were paired when their similarity score exceeded a predefined threshold, ensuring high-quality matches.

One key challenge was the scarcity of point clouds with corresponding point-of-view (POV) images. While datasets like \textbf{KITTI} provide full 360-degree point clouds, their associated images typically capture only limited portions of the scene, leading to incomplete visual coverage. This mismatch makes such datasets unsuitable for high-resolution reconstruction tasks that rely on aligning image cues with geometric structures. \textbf{ObitoNet} specifically depends on aligned POV point clouds and images to learn how to fill in gaps and refine details, a capability not achievable with datasets that lack comprehensive correspondence.

\subsubsection{Point Cloud Data} The {\textit{Tanks and Temples}} dataset includes dense and detailed 3D point clouds reconstructed from multi-view stereo (MVS) algorithms. These point clouds represent complex objects and scenes such as buildings, statues, and landscapes. The raw point cloud data is processed as follows:

\begin{itemize} 
\item \textbf{Normalization:} All point clouds are normalized to fit within a unit sphere centered at the origin to ensure scale invariance during training.
\item \textbf{Downsampling:} To reduce computational overhead, point clouds are downsampled using Furthest Point Sampling (FPS), which retains the most geometrically representative points.
\item \textbf{Grouping:} FPS-sampled centers serve as anchors, and a K-Nearest Neighbors (KNN) algorithm groups neighboring points into clusters of size \textit{M}. This results in structured input groups $\mathbf{N} \in \mathbb{R}^{B \times G \times M \times 3}$, where \textit{B} is the batch size, \textit{G} is the number of tokens, and \textit{M} is the number of points in each group. \end{itemize}

The resulting point cloud tokens $\mathbf{P} \in \mathbb{R}^{B \times G \times C}$ are extracted using the PCTokenizer module and passed to the transformer encoder.

\subsubsection{Image Data} The dataset also provides corresponding multi-view RGB images, ensuring spatial alignment with the 3D point clouds. Each image captures the scene from a unique camera perspective, ensuring full coverage of the target object or scene. The image preprocessing steps are as follows:

\begin{itemize} 
\item \textbf{Resizing:} All images are resized to a fixed resolution \textit{H×W} to ensure compatibility with the Vision Transformer (ViT).
\item \textbf{Patch Embedding:} Each image is divided into non-overlapping patches of size \textit{P×P}, where \textit{P} is determined based on the number of point cloud groups \textit{G}. The patches are flattened and projected into a latent embedding space. 
\item \textbf{Feature Extraction:} 
The images are passed through a pretrained Vision Transformer (ViT). The CLS token is discarded, and the remaining patch tokens $\mathbf{T}_{img} \in \mathbb{R}^{B \times G \times C}$ are retained as input tokens for cross-modal fusion.
\end{itemize}

\subsection{Multimodal Fusion via Cross-Attention} 
The Cross-Attention mechanism acts as a bridge between the two modalities—image tokens and point cloud tokens—facilitating effective information exchange and alignment. The point cloud tokens encode compact geometric features through grouping and sampling, while the image tokens, extracted by the \textbf{Vision Transformer (ViT)}, provide rich semantic representations by processing the input image as sequences of patches. The encoded point cloud tokens, and the image tokens are seamlessly fused through the Cross-Attention mechanism, enabling the model to leverage complementary features for enhanced reconstruction.

Within the \textbf{Cross-Attention block}, the point cloud tokens serve as the primary focus, acting as the \textit{queries} (\textbf{Q}), while the image tokens provide complementary information as the \textit{keys} (\textbf{K}) and \textit{values} (\textbf{V}). This interaction enables the model to establish meaningful relationships between the two modalities. Specifically, the point cloud tokens query the image tokens to extract relevant semantic information that aligns with and enhances the geometric structures represented in the point cloud.

\subsection{Transformer-Based Decoder} 
The fused features obtained from the Cross-Attention block are further processed by a \textbf{Transformer-based Decoder}, which refines these features to ensure they are coherent, detailed, and ready for reconstruction. The decoder consists of multiple layers that iteratively process the fused features using self-attention and feedforward networks, enabling the model to capture fine-grained details and improve feature quality.

Within the Transformer, Self-Attention ensures that each token interacts with all other tokens in the feature set, enabling the model to capture and refine both global and local relationships between geometric and semantic information. To stabilize the learning process and maintain effective gradient flow during training, residual connections are employed. 

After the decoder refines the fused features, the reconstruction head transforms these features into a set of 3D points generating the point cloud. The reconstruction head employs a sequential architecture consisting of two 1D convolutional layers and a ReLU activation function. The first convolutional layer increases the feature dimensionality, while the ReLU activation introduces non-linearity to capture complex relationships. The second convolutional layer maps the features to the final 3D coordinates of the point cloud, producing output dimensions aligned with the target group size.

\subsubsection{Chamfer Loss}
Chamfer Loss evaluates the distance between the predicted point cloud and the ground truth point cloud by measuring how close the points in one set are to their nearest neighbors in the other set. This property makes it particularly effective for tasks such as point cloud reconstruction and completion.

To evaluate the alignment between the predicted point cloud \( P_{\text{recon}} \) and the ground truth point cloud \( P_{\text{gt}} \), we utilize the Chamfer Loss, defined as:

\begin{equation}
\mathcal{L}_{\text{Chamfer}}(P_{\text{recon}}, P_{\text{gt}}) = \frac{1}{|P_{\text{recon}}|} \sum_{x \in P_{\text{recon}}} \min_{y \in P_{\text{gt}}} \|x - y\|_2^2 
+ \frac{1}{|P_{\text{gt}}|} \sum_{y \in P_{\text{gt}}} \min_{x \in P_{\text{recon}}} \|y - x\|_2^2,
\label{eq:chamfer_loss}
\end{equation}

where \( \| \cdot \|_2 \) denotes the Euclidean distance, and \( |\cdot| \) refers to the number of points in each set.

\subsection{Train Order}
The pretrained transformers utilized in our framework have been primarily trained for tasks such as classification and segmentation. Consequently, a carefully defined training sequence was required to ensure that each model could function as an independent unit while contributing to the overall reconstruction task. The training process was divided into three distinct stages:

\begin{itemize}
\item \textbf{Train Order 1:} In this stage, we trained \textbf{ObitoNetPC} and \textbf{ObitoNetCA} independently, replicating the approach of the \textbf{PointMAE} [3] architecture. The objective here was to train the model to fill gaps in the point cloud using only point cloud tokens.

\item \textbf{Train Order 2:} In the second stage, the \textbf{ObitoNetPC} and \textbf{ObitoNetCA} models were frozen, and the Cross-Attention module within \textbf{ObitoNetCA} was activated. This stage preserved the knowledge learned by ObitoNetPC and ObitoNetCA while allowing ObitoNetImg to generate image tokens specifically optimized for the reconstruction task.

\item \textbf{Train Order 3:} In the final stage, all three models were trained together to perform the reconstruction task. This step allowed the models to collaboratively learn features specific to their tasks while refining their outputs.
\end{itemize}

This sequential training approach ensured that each model contributed effectively to the reconstruction pipeline, while also allowing the models to integrate their respective modalities seamlessly.

\section{Experiments}
As outlined in the Dataset Generation section, we use the {\textit{Tanks and Temples}} Dataset, which includes point clouds and corresponding images, to create point cloud-image pairs for filling gaps in the point cloud. The model was trained with these pairs by providing a masked point cloud and a scene image as inputs. The model was tasked with reconstructing a masked point cloud with three times the input points, upsampling and restoring missing geometry. Training minimized the Chamfer Distance loss between the reconstructed and original point clouds, enabling accurate restoration of fine-grained details while maintaining geometric consistency.

\subsection{Experimental Setup}
The experiments were conducted on 4 NVIDIA RTX A5000 GPUs within the Nexus cluster provided by the University of Maryland. The specific parameters and architectures used by each model are detailed in their respective subsections for clarity and reproducibility. The computational setup ensured efficient training and evaluation of the proposed framework.

\subsection{ObitoNet/Base Model}

\begin{figure}[H]
    \centering

    \begin{subfigure}[b]{0.55\textwidth} 
        \centering
        \includegraphics[width=\textwidth]{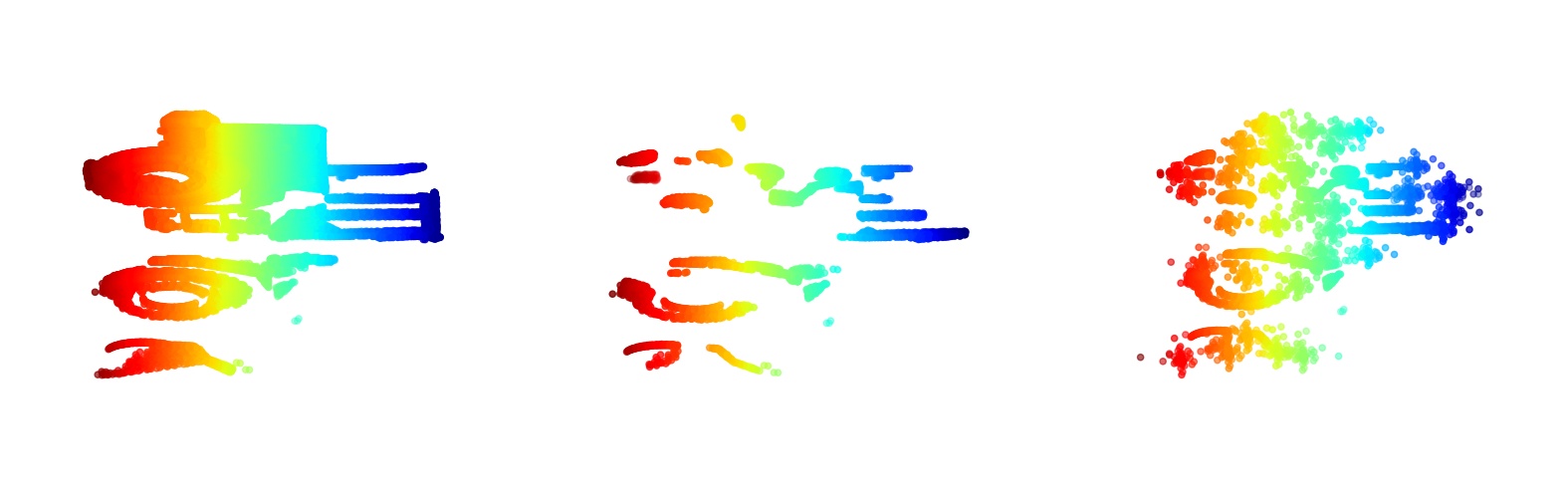}
        \vspace{0.5em} 
        \includegraphics[width=\textwidth]{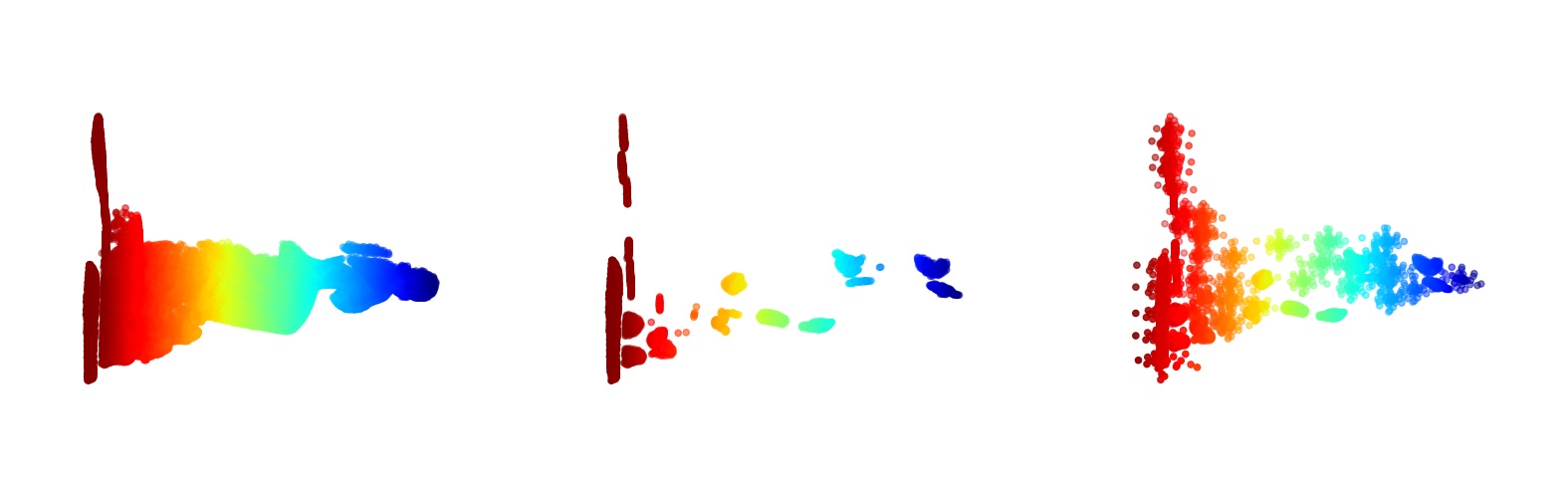}
        \caption{Point Cloud Visualizations}
        \label{fig:pointclouds(2)}
    \end{subfigure}
    \hfill
    \begin{subfigure}[b]{0.4\textwidth} 
        \centering
        \includegraphics[width=\textwidth, height=0.45\textwidth, keepaspectratio]{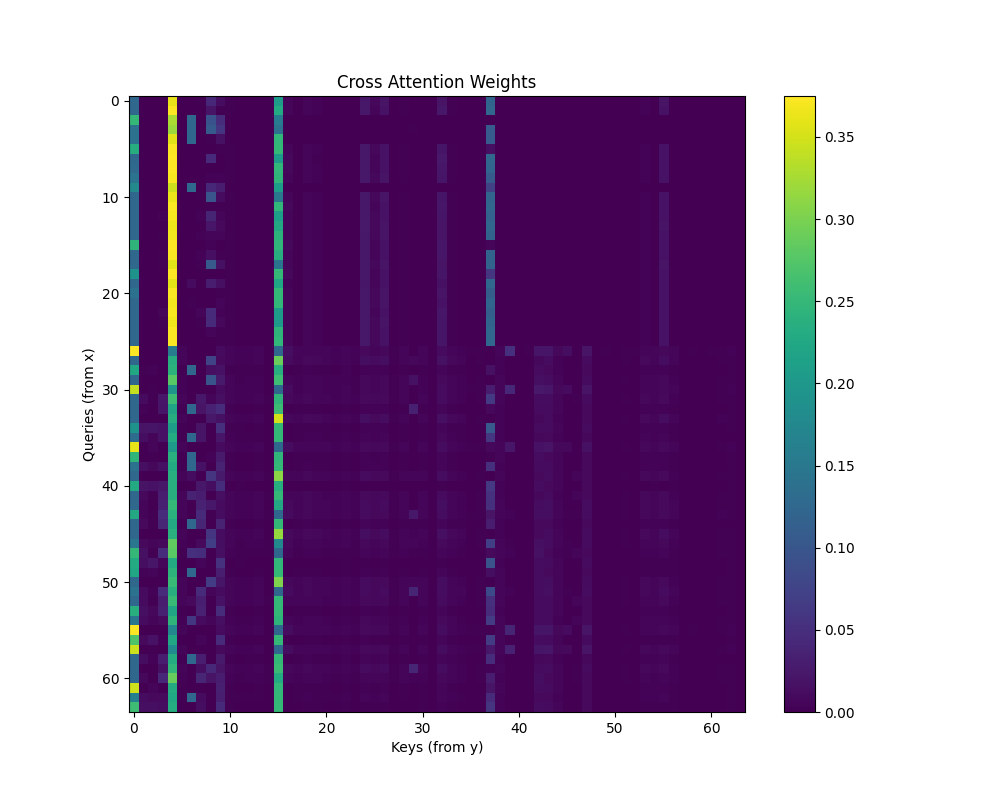}
        \vspace{0.5em} 
        \includegraphics[width=\textwidth, height=0.45\textwidth, keepaspectratio]{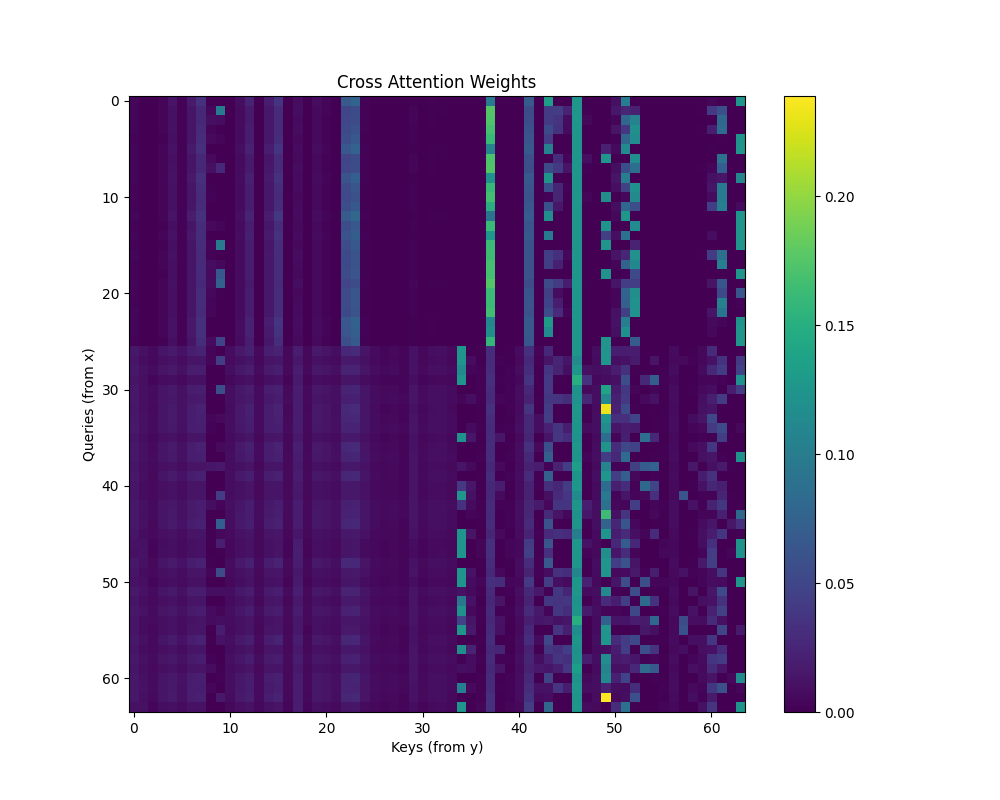}
        \caption{Cross Attention Weights}
        \label{fig:cross_attention(2)}
    \end{subfigure}

    \captionsetup{justification=centering}
    \caption{Side-by-side comparison of (a) Point Cloud Visualizations and (b) Cross Attention Weights for the ObitoNet/Base Model}
    \label{fig:combined_visualization(1)}
\end{figure}

The base \textbf{ObitoNet} model was trained using the specified \textbf{Train Order sequence}, ensuring a structured approach to model development. Table 1 highlights the key details of the model configuration along with the sample outputs in Figure 3

\begin{table}[ht!]
\centering
\renewcommand{\arraystretch}{1.3} 
\begin{tabular}{|l|l|}
\hline
\textbf{Parameter}                  & \textbf{Value}         \\ \hline
\textbf{Token Size}                 & 768                   \\ \hline
\textbf{Number of Tokens}           & 64                    \\ \hline
\textbf{MAE Decoder Depth}          & 4                     \\ \hline
\textbf{Vision Transformer (ViT) Model} & Base Google ViT Model \\ \hline
\end{tabular}
\vspace{5pt} 
\caption{Model Configuration Details}
\label{table:model_details}
\end{table}

\subsection{ObitoNet/ViTMAE Model}
The \textbf{ViTMAE} model shares the same architecture as the \textbf{ViT Base} model, with the key difference being the use of \textbf{ViTMAE} as the image encoder instead of the standard \textbf{ViT}. 
The choice to use \textbf{ViTMAE} over \textbf{ViT} stems from its suitability for reconstruction tasks involving hidden input point clouds. Unlike standard ViT, ViTMAE leverages self-supervised pretraining to reconstruct missing image patches, making it highly effective for tasks that require inferring and filling gaps. This aligns naturally with point cloud reconstruction, where the model needs to restore missing geometric details. By leveraging robust semantic features from images, ViTMAE enables more accurate and context-aware reconstructions, making it a superior choice for this task.

\begin{figure}[H]
    \centering

    \begin{subfigure}[b]{0.55\textwidth} 
        \centering
        \includegraphics[width=\textwidth]{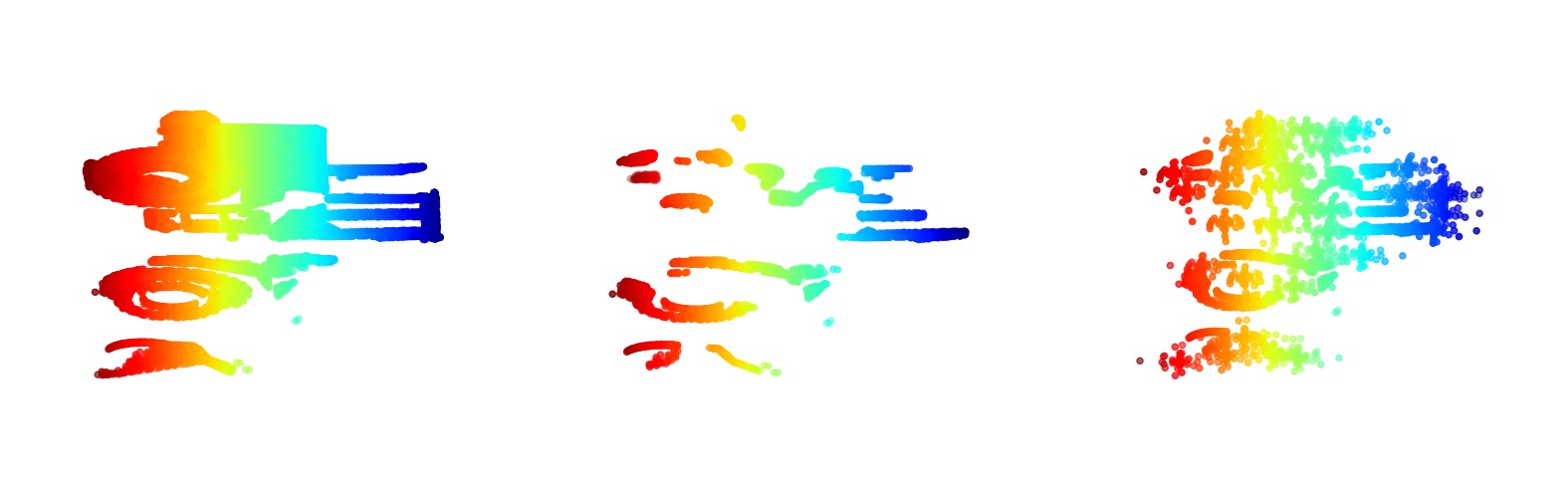}
        \vspace{0.5em} 
        \includegraphics[width=\textwidth]{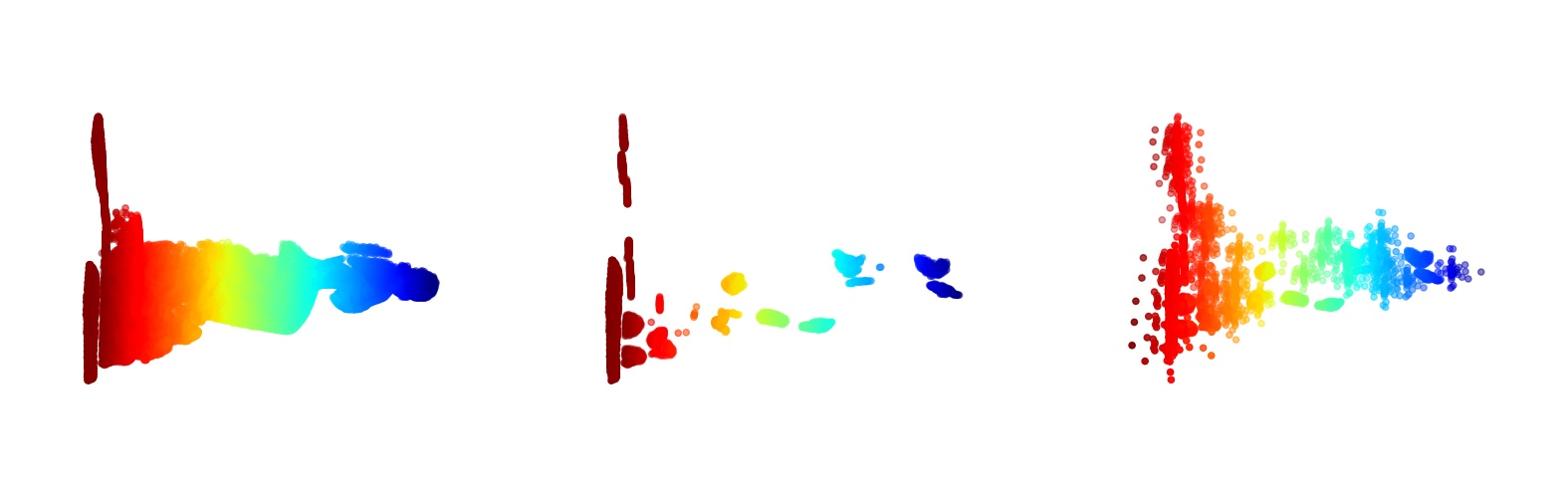}
        \caption{Point Cloud Visualizations}
        \label{fig:pointclouds(3)}
    \end{subfigure}
    \hfill
    \begin{subfigure}[b]{0.4\textwidth} 
        \centering
        \includegraphics[width=\textwidth, height=0.45\textwidth, keepaspectratio]{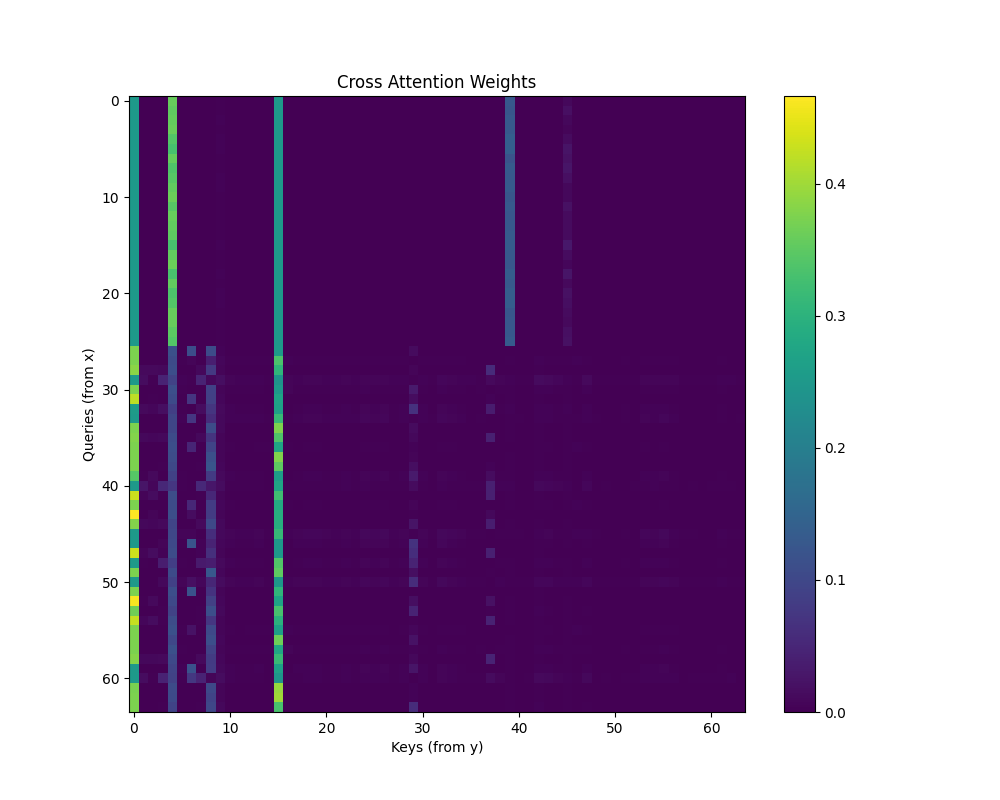}
        \vspace{0.5em} 
        \includegraphics[width=\textwidth, height=0.45\textwidth, keepaspectratio]{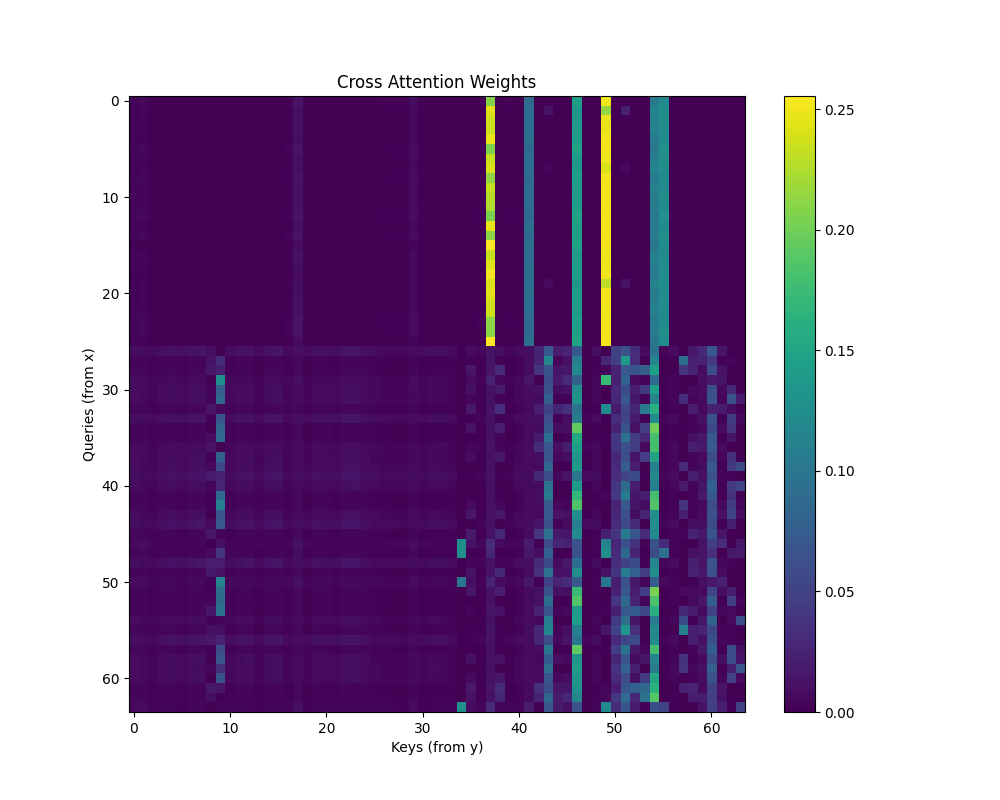}
        \caption{Cross Attention Weights}
        \label{fig:cross_attention(3)}
    \end{subfigure}

    \captionsetup{justification=centering}
    \caption{Side-by-side comparison of (a) Point Cloud Visualizations and (b) Cross Attention Weights for the ObitoNet/ViTMAE Model}
    \label{fig:combined_visualization(2)}
\end{figure}

\begin{table}[ht!]
\centering
\renewcommand{\arraystretch}{1.3} 
\begin{tabular}{|l|l|}
\hline
\textbf{Parameter}                  & \textbf{Value}         \\ \hline
\textbf{Token Size}                 & 768                    \\ \hline
\textbf{Number of Tokens}           & 64                     \\ \hline
\textbf{MAE Decoder Depth}          & 4                      \\ \hline
\textbf{Vision Transformer (ViT) Model} & ViTMAE Model        \\ \hline
\end{tabular}
\vspace{5pt} 
\caption{Model Configuration Details for ObitoNet/ViTMAE Variant}
\label{table:model_vitmae_details}
\end{table}

The results of this experiment are highlighted in Table 2, with sample outputs in Figure 4

\subsection{ObitoNet/Large Model}
The \textbf{ObitoNet/Large} model was designed to evaluate the scalability and effectiveness of the model with a larger token set and deeper decoder architecture. In this configuration, the \textbf{ViT} model was used as the image tokenizer, with an increased \textbf{number of tokens} and \textbf{MAE decoder depth} to enhance the capacity for learning complex features. This model was trained end-to-end for 2000 epochs to assess its performance under a prolonged training regime.

\begin{figure}[ht]
    \centering

    \begin{subfigure}[b]{0.55\textwidth} 
        \centering
        \includegraphics[width=\textwidth]{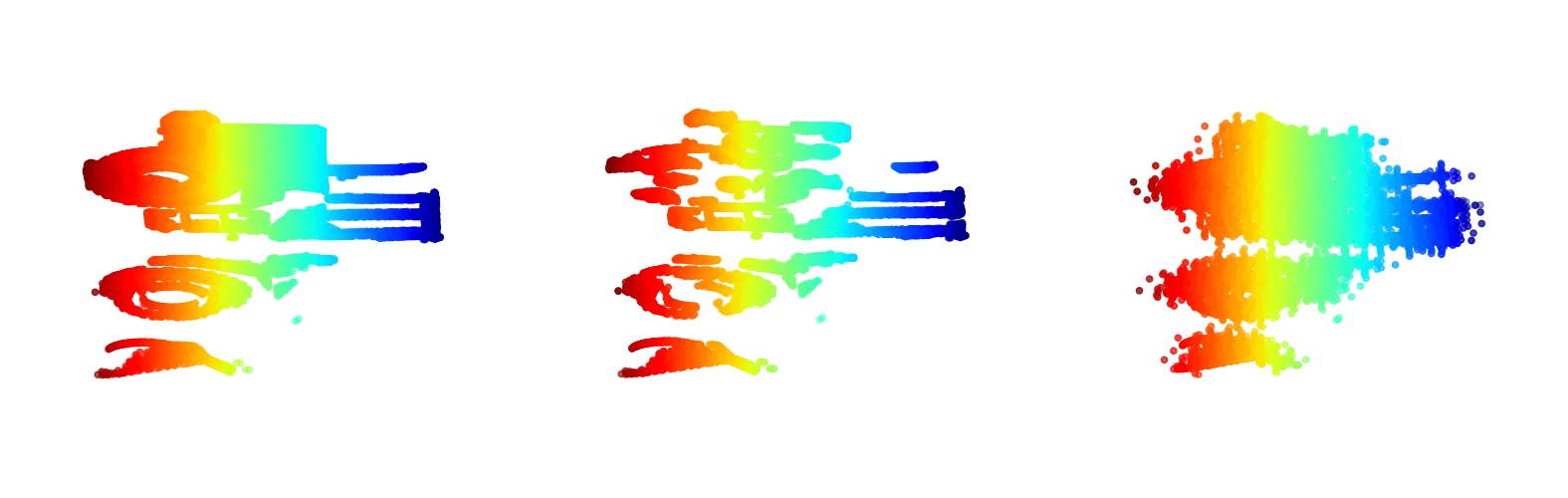}
        \vspace{0.5em} 
        \includegraphics[width=\textwidth]{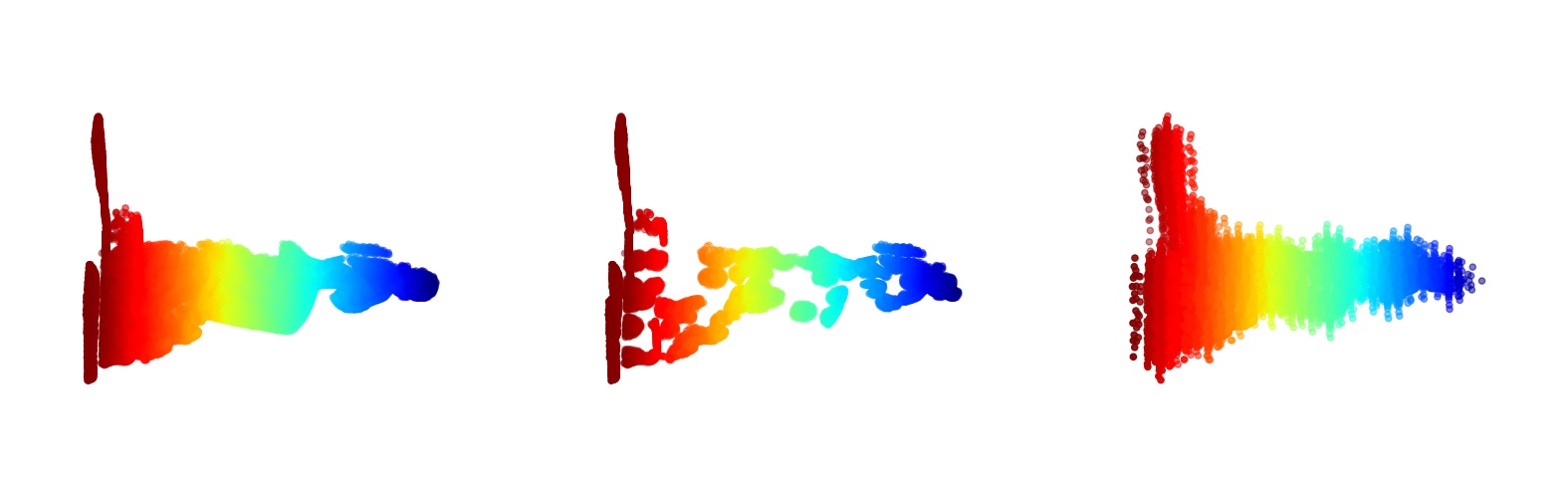}
        \caption{Point Cloud Visualizations}
        \label{fig:pointclouds(4)}
    \end{subfigure}
    \hfill
    \begin{subfigure}[b]{0.4\textwidth} 
        \centering
        \includegraphics[width=\textwidth, height=0.45\textwidth, keepaspectratio]{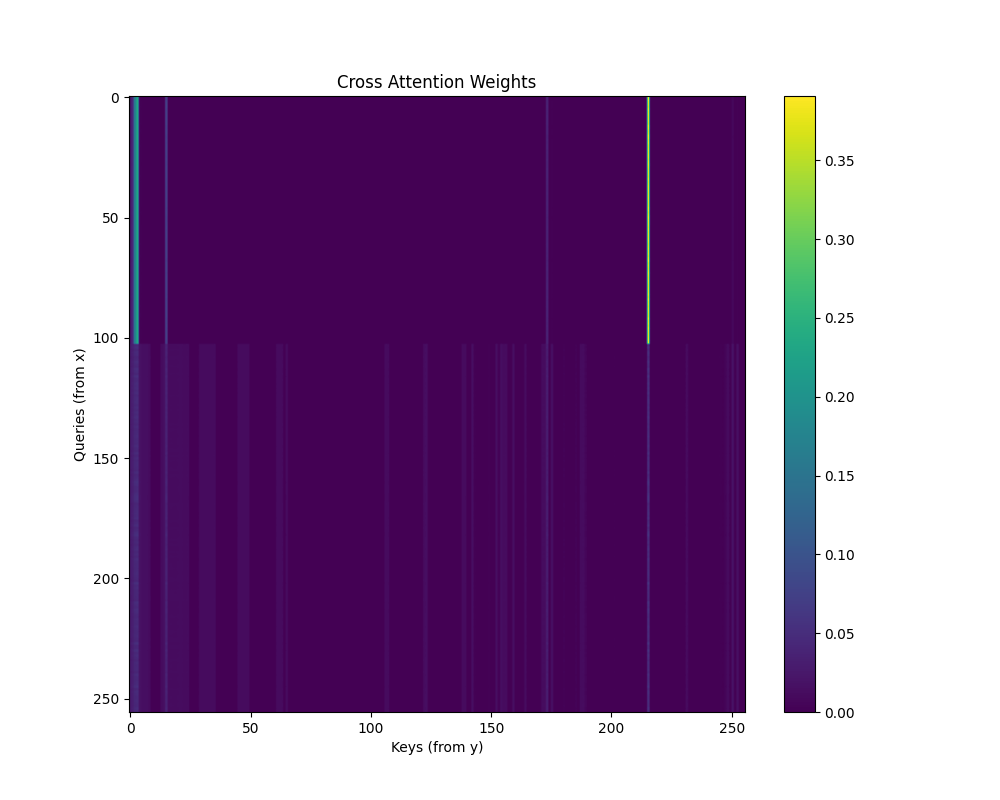}
        \vspace{0.5em} 
        \includegraphics[width=\textwidth, height=0.45\textwidth, keepaspectratio]{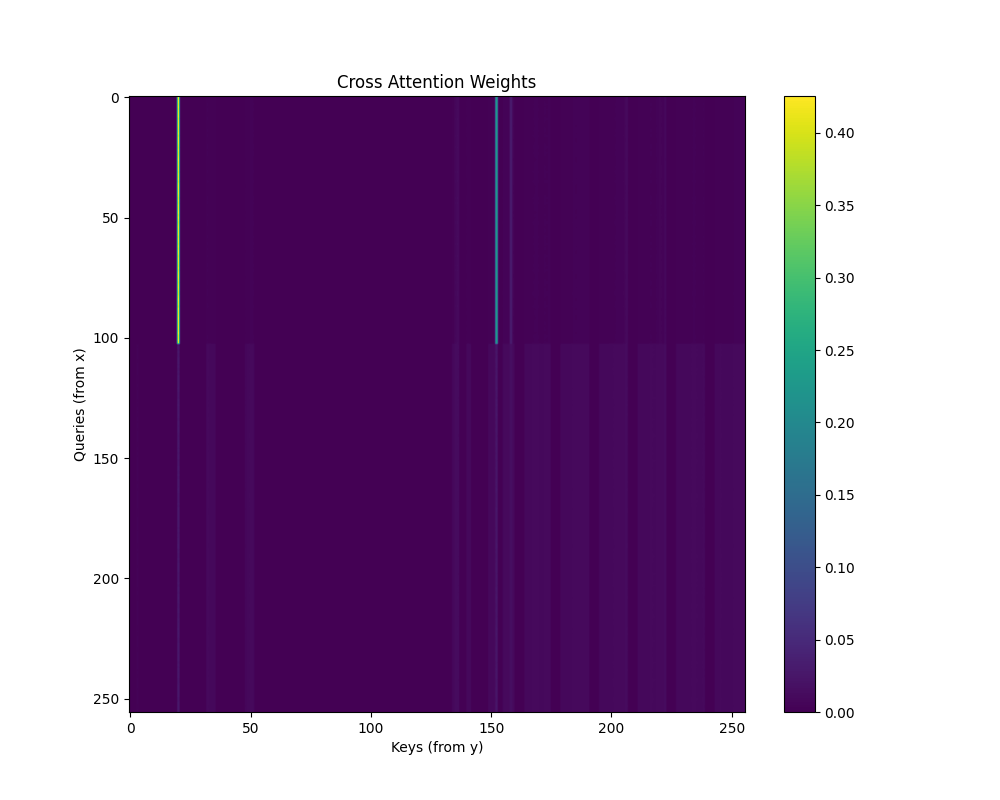}
        \caption{Cross Attention Weights}
        \label{fig:cross_attention(1)}
    \end{subfigure}

    \captionsetup{justification=centering}
    \caption{Side-by-side comparison of (a) Point Cloud Visualizations and (b) Cross Attention Weights for the ObitoNet/Large Model}
    \label{fig:combined_visualization(3)}
\end{figure}

\begin{table}[ht!]
\centering
\renewcommand{\arraystretch}{1.3} 
\begin{tabular}{|l|l|}
\hline
\textbf{Parameter}                  & \textbf{Value}         \\ \hline
\textbf{Token Size}                 & 768                   \\ \hline
\textbf{Number of Tokens}           & 256                   \\ \hline
\textbf{MAE Decoder Depth}          & 12                    \\ \hline
\textbf{Vision Transformer (ViT) Model} & Base Google ViT Model  \\ \hline
\end{tabular}
\vspace{5pt}
\caption{Model Configuration Details for ObitoNet/Large Variant}
\label{table:model_large_details}
\end{table}

In this experiment, the model’s larger capacity and deeper architecture were tested for their ability to handle high-resolution point cloud reconstruction and multimodal integration. The results highlight the trade-offs between model size, training duration, and performance, offering valuable insights into scaling strategies for multimodal frameworks.

\begin{table}[ht!]
\centering
\renewcommand{\arraystretch}{1.3} 
\begin{tabular}{|l|c|}
\hline
\textbf{Model}                     & \textbf{Chamfer Loss} \\ \hline
\textbf{ObitoNet/Base}             & 1.36               \\ \hline
\textbf{ObitoNet/Large}            & 1.20                \\ \hline
\textbf{ObitoNet/ViTMAE}           & 1.41               \\ \hline
\textbf{PointMAE}     & 1.53                \\ \hline
\end{tabular}
\vspace{5pt} 
\caption{Chamfer Loss values for different models.}
\label{table:chamfer_loss}
\end{table}

\section{Conclusion}
We proposed a Cross-Attention-based pipeline for high-resolution point cloud reconstruction. By integrating Vision Transformer image features with KNN-based point cloud processing, we achieve results comparable to PointMAE. The modular design of ObitoNet, consisting of three distinct components—the image tokenizer (\textbf{ObitoNetImg}), the point cloud tokenizer (\textbf{ObitoNetPC}), and the Cross-Attention module—opens opportunities for further exploration and downstream tasks. The image tokenizer, built on the \textbf{ViT} or \textbf{ViTMAE} architectures, extracts semantic features, making it suitable for tasks such as image classification, object detection, and semantic segmentation. Similarly, the point cloud tokenizer, utilizing FPS and KNN, captures compact geometric features that can be adapted for 3D object classification, scene understanding, and point cloud registration. Beyond these individual components, the Cross-Attention module offers potential for multimodal applications such as augmented reality, robotics, and spatial reasoning tasks, where fusing 2D and 3D data is essential. By fine-tuning these tokenizers for specific use cases, ObitoNet can serve as a versatile framework to address diverse challenges in computer vision and 3D processing, contributing to advancements in multimodal learning.

\setlength{\bibsep}{1pt}
\section{Appendix}

\subsection{ObitoNet/Base}
\begin{figure}[H]
    \centering

    \begin{subfigure}[t]{0.6\textwidth} 
        \centering
        \includegraphics[width=\textwidth]{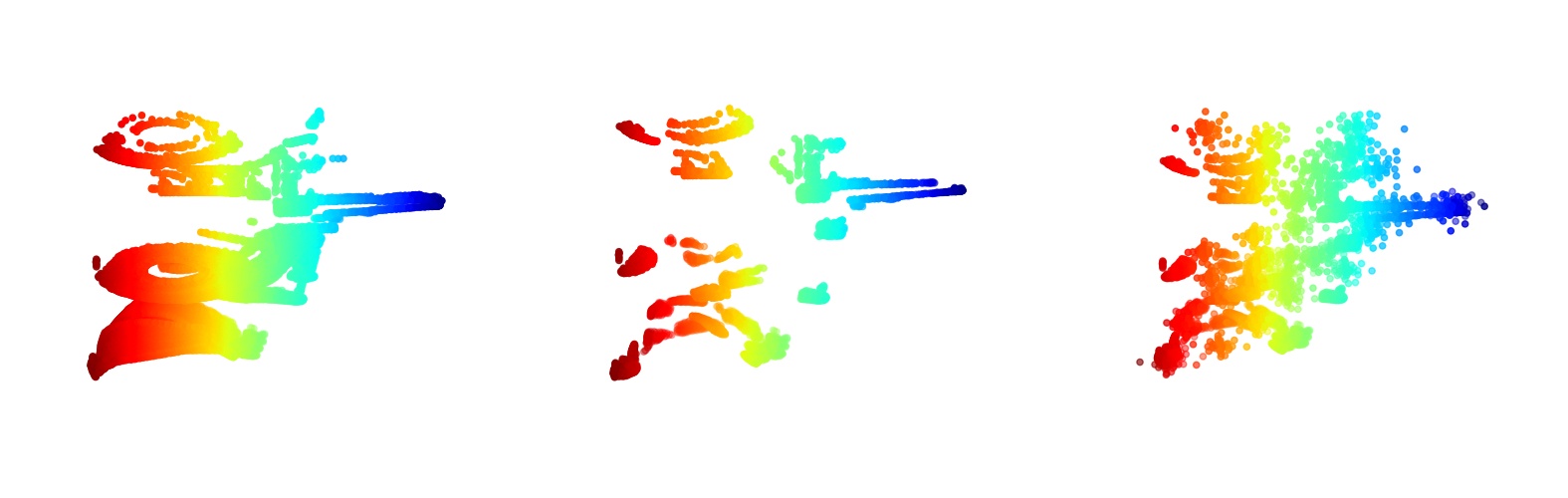}
    \end{subfigure}
    \hfill
    \begin{subfigure}[t]{0.38\textwidth} 
        \centering
        \includegraphics[width=\textwidth, height=0.6\textwidth, keepaspectratio]{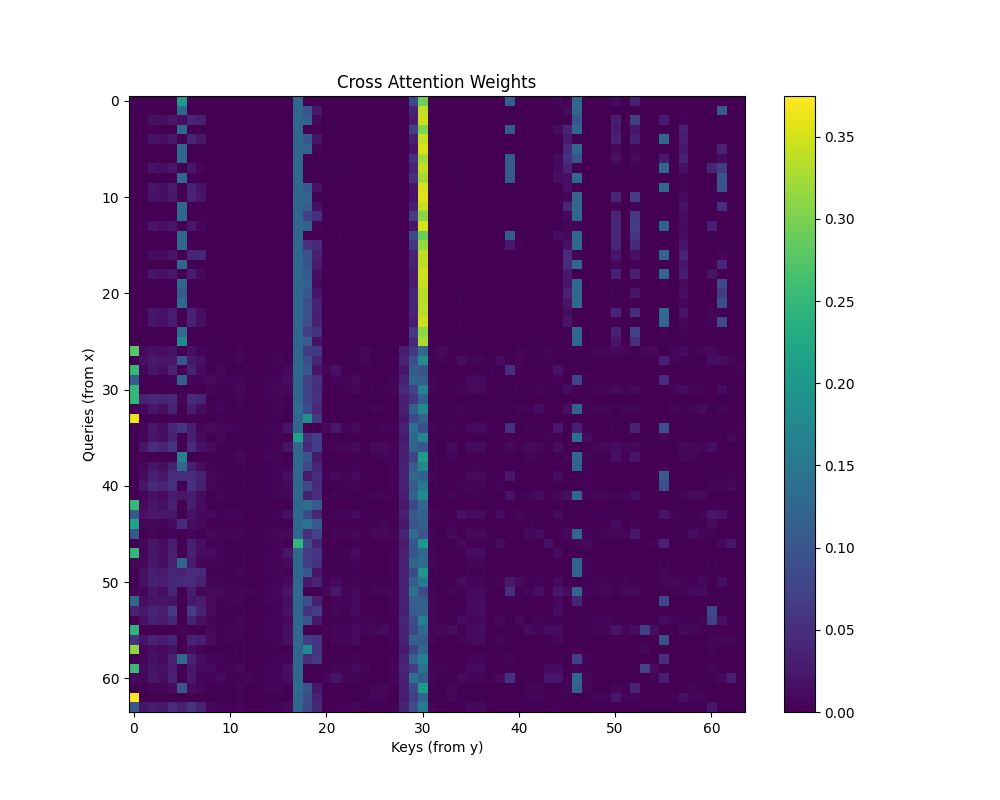}
    \end{subfigure}

    \vspace{1em}

    \begin{subfigure}[t]{0.6\textwidth} 
        \centering
        \includegraphics[width=\textwidth]{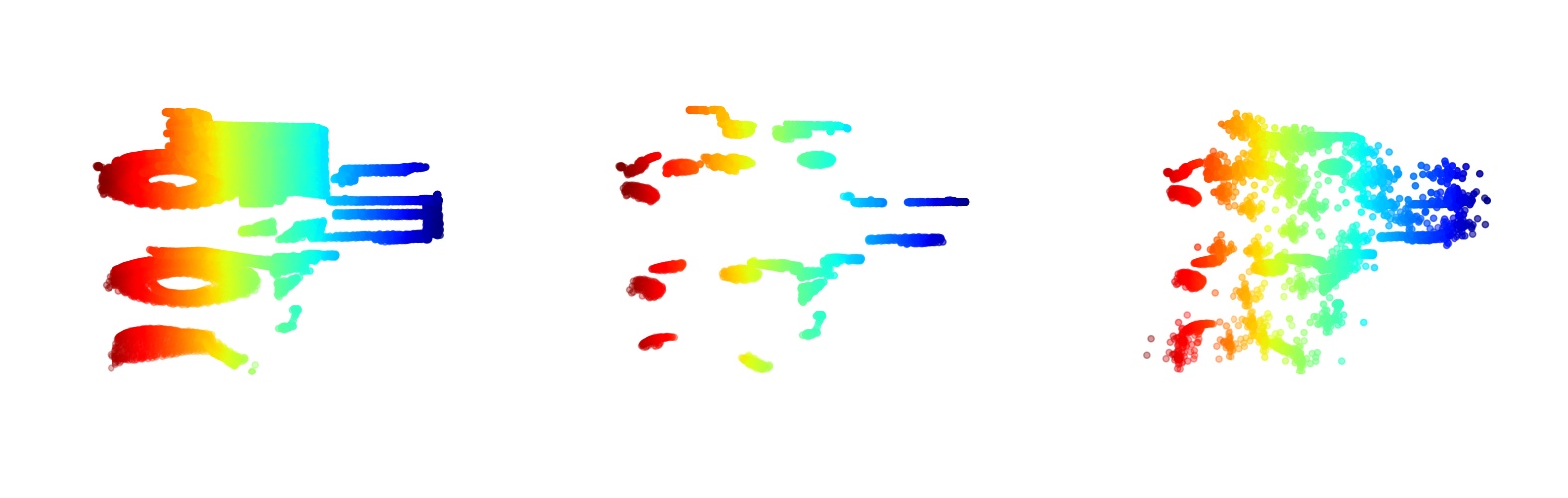}
    \end{subfigure}
    \hfill
    \begin{subfigure}[t]{0.38\textwidth} 
        \centering
        \includegraphics[width=\textwidth, height=0.6\textwidth, keepaspectratio]{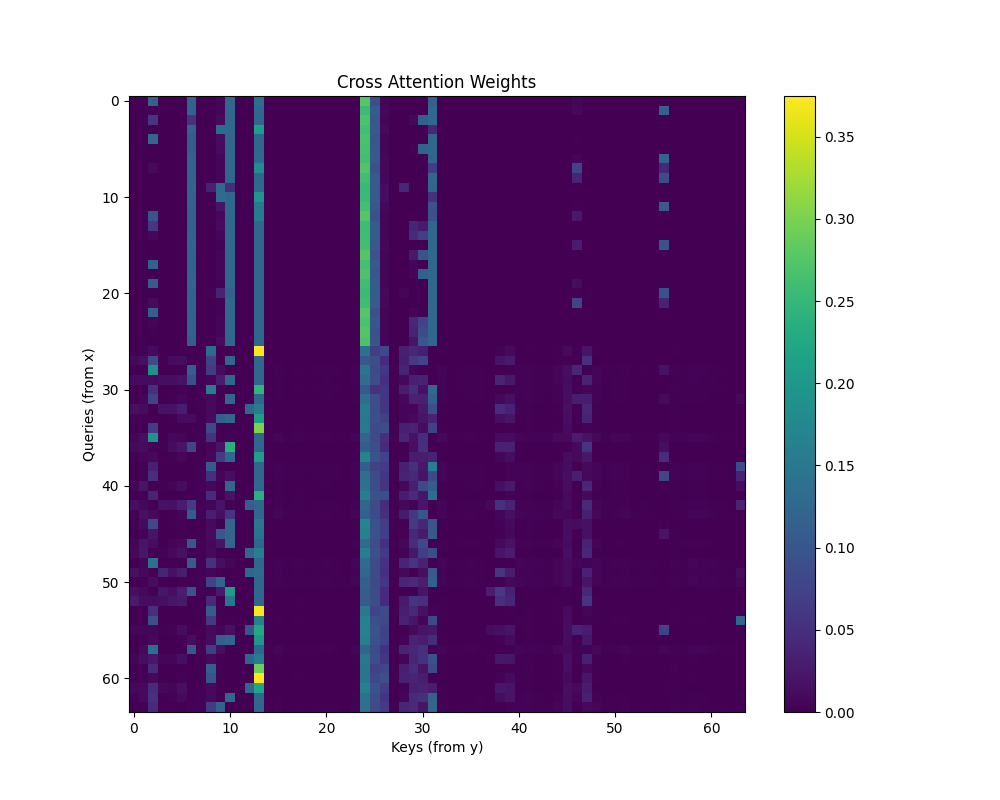}
    \end{subfigure}

    \vspace{1em}

    \begin{subfigure}[t]{0.6\textwidth} 
        \centering
        \includegraphics[width=\textwidth]{base_plot_0035.jpg}
    \end{subfigure}
    \hfill
    \begin{subfigure}[t]{0.38\textwidth} 
        \centering
        \includegraphics[width=\textwidth, height=0.6\textwidth, keepaspectratio]{base_attention_0035.png}
    \end{subfigure}

    \vspace{1em}

    \begin{subfigure}[t]{0.6\textwidth} 
        \centering
        \includegraphics[width=\textwidth]{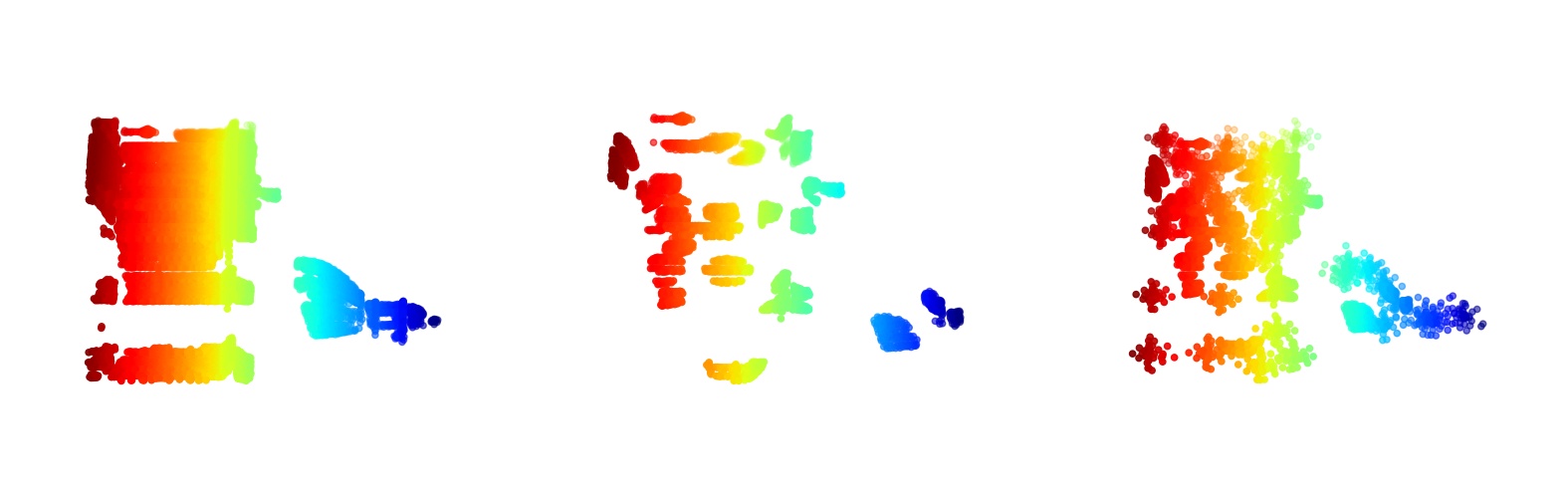}
    \end{subfigure}
    \hfill
    \begin{subfigure}[t]{0.38\textwidth} 
        \centering
        \includegraphics[width=\textwidth, height=0.6\textwidth, keepaspectratio]{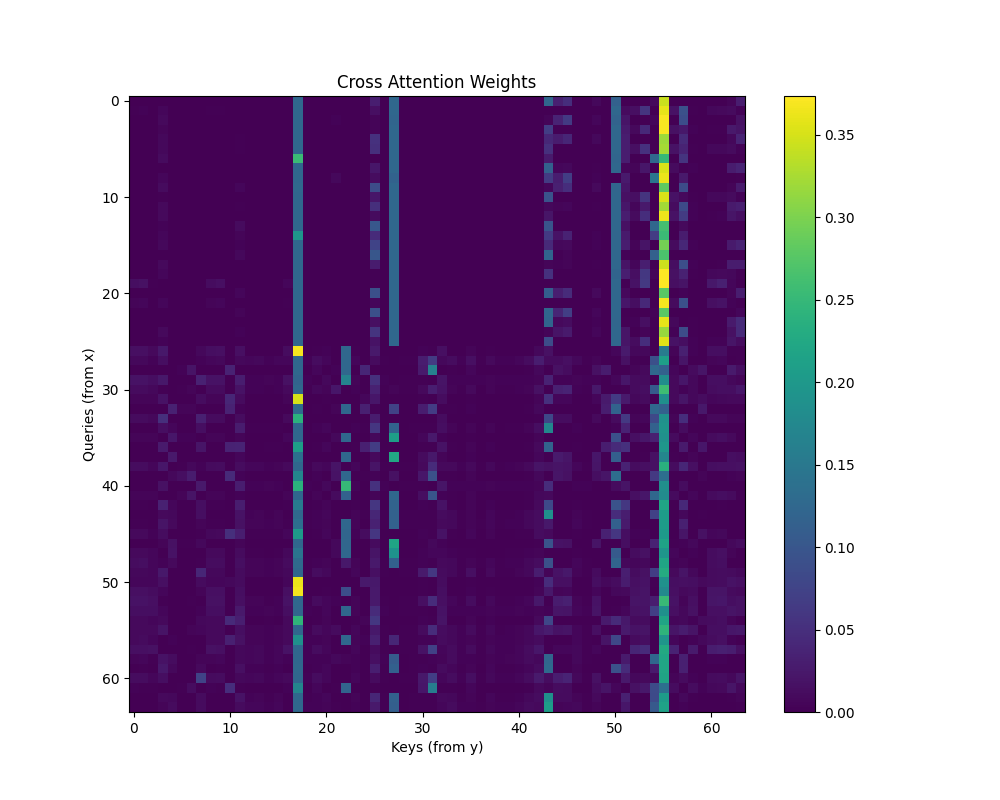}
    \end{subfigure}

    \vspace{1em}

    \begin{subfigure}[t]{0.6\textwidth} 
        \centering
        \includegraphics[width=\textwidth]{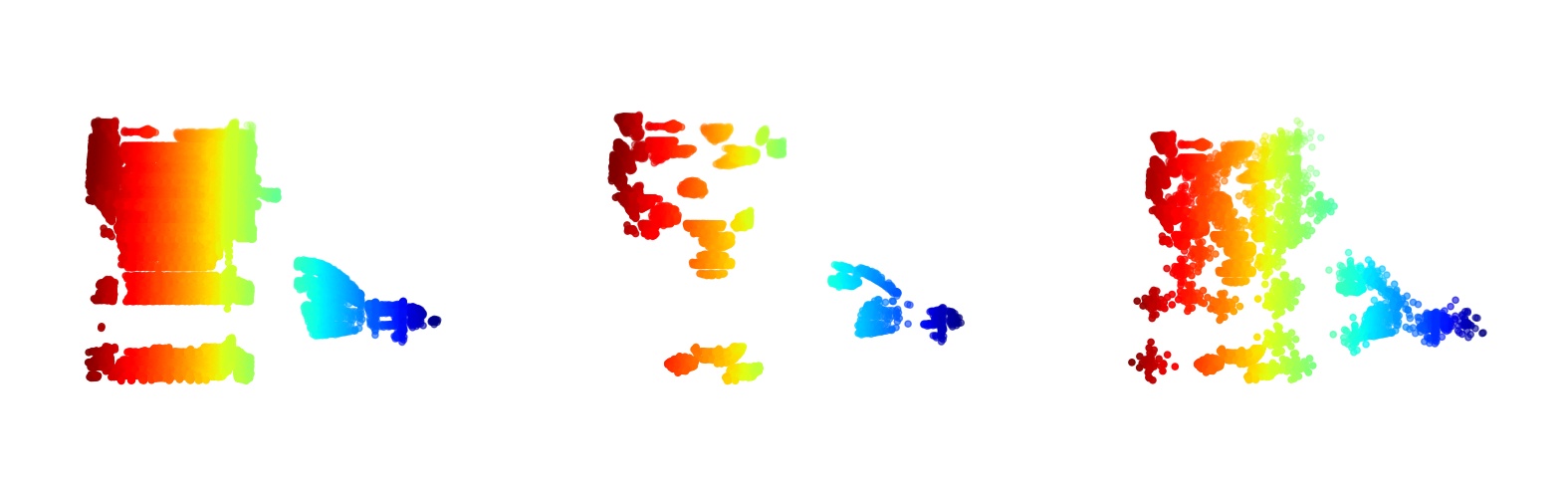}
    \end{subfigure}
    \hfill
    \begin{subfigure}[t]{0.38\textwidth} 
        \centering
        \includegraphics[width=\textwidth, height=0.6\textwidth, keepaspectratio]{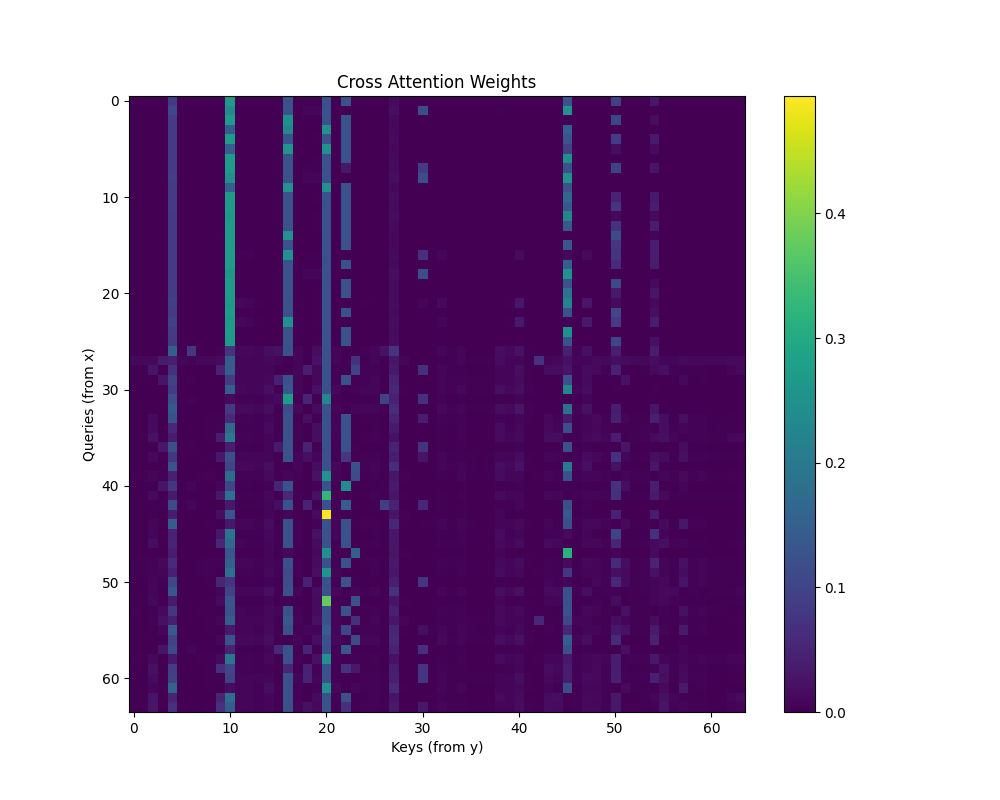}
    \end{subfigure}

    \vspace{1em}

    \begin{subfigure}[t]{0.6\textwidth} 
        \centering
        \includegraphics[width=\textwidth]{base_plot_0088.jpg}
    \end{subfigure}
    \hfill
    \begin{subfigure}[t]{0.38\textwidth} 
        \centering
        \includegraphics[width=\textwidth, height=0.6\textwidth, keepaspectratio]{base_attention_0088.png}
    \end{subfigure}

    \caption{Further ObitoNet/Base model sample outputs.}
    \label{fig:obitonet_base_outputs}
\end{figure}

\subsection{ObitoNet/ViTMAE}
\begin{figure}[H]
    \centering

    \begin{subfigure}[t]{0.6\textwidth} 
        \centering
        \includegraphics[width=\textwidth]{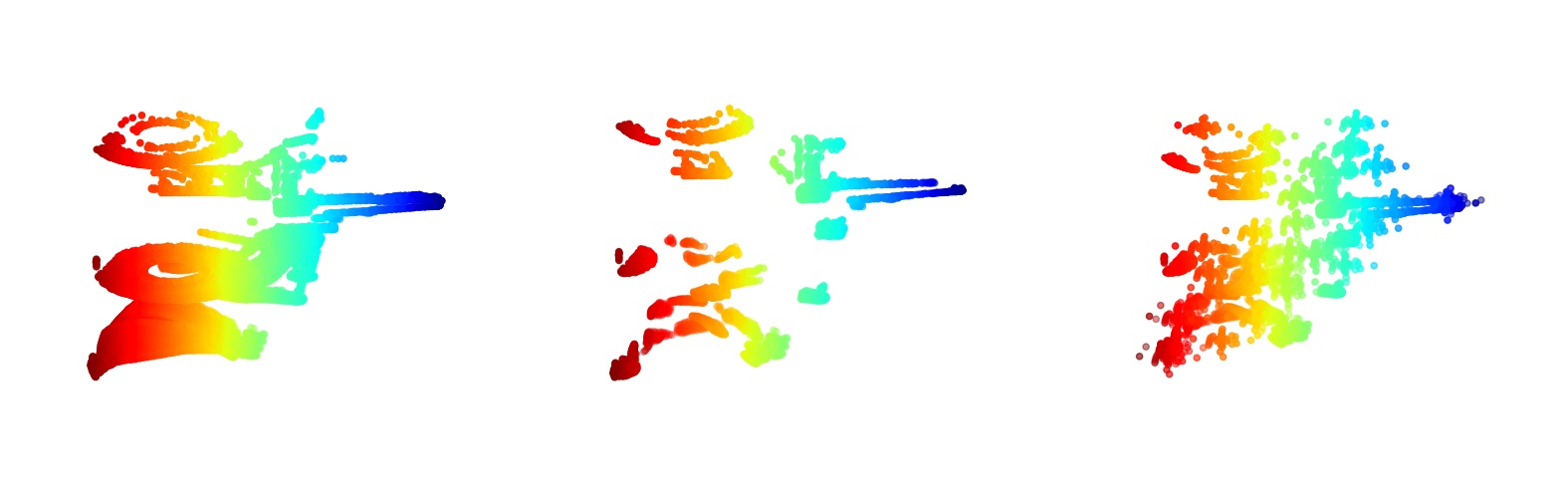}
    \end{subfigure}
    \hfill
    \begin{subfigure}[t]{0.38\textwidth} 
        \centering
        \includegraphics[width=\textwidth, height=0.6\textwidth, keepaspectratio]{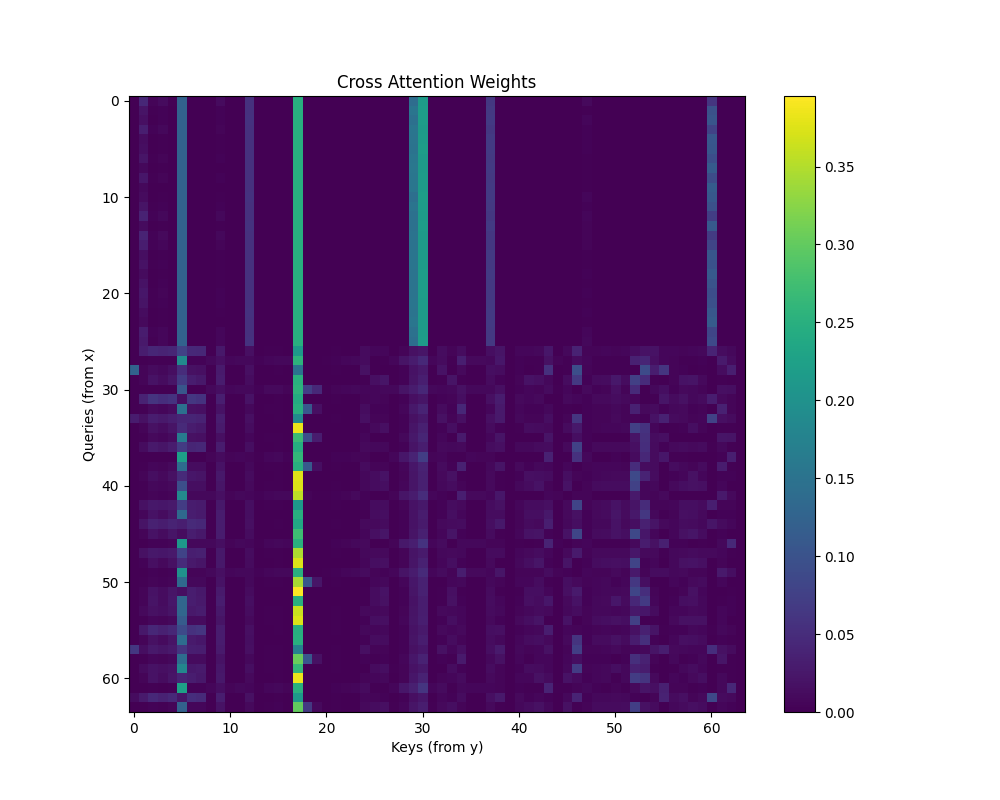}
    \end{subfigure}

    \vspace{1em}

    \begin{subfigure}[t]{0.6\textwidth} 
        \centering
        \includegraphics[width=\textwidth]{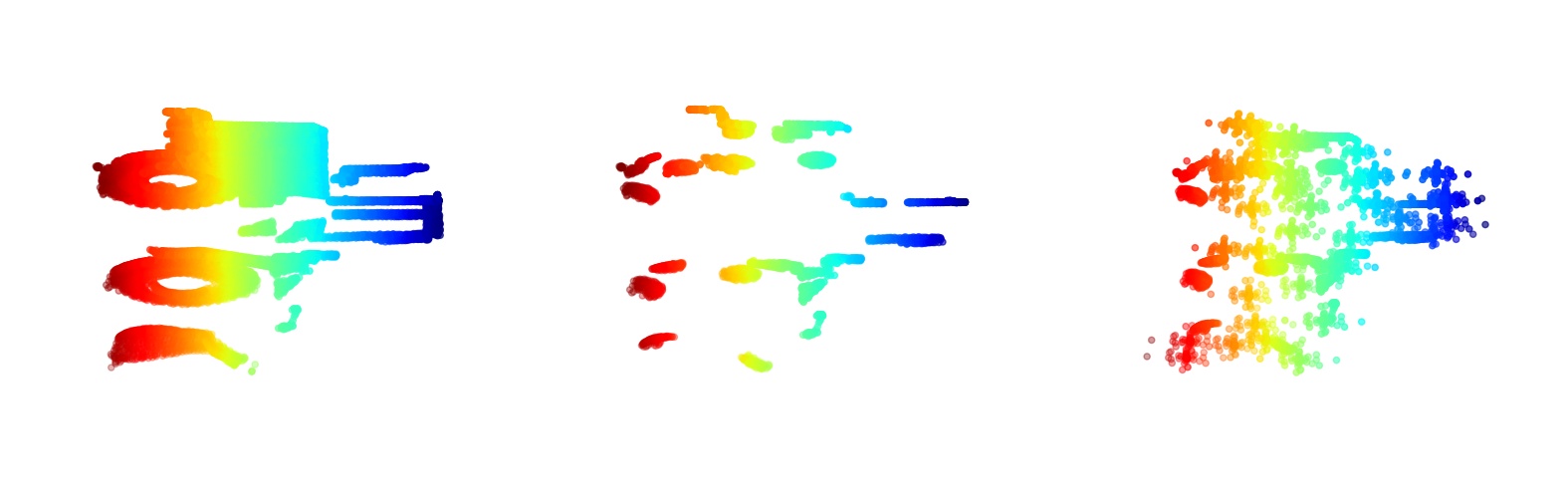}
    \end{subfigure}
    \hfill
    \begin{subfigure}[t]{0.38\textwidth} 
        \centering
        \includegraphics[width=\textwidth, height=0.6\textwidth, keepaspectratio]{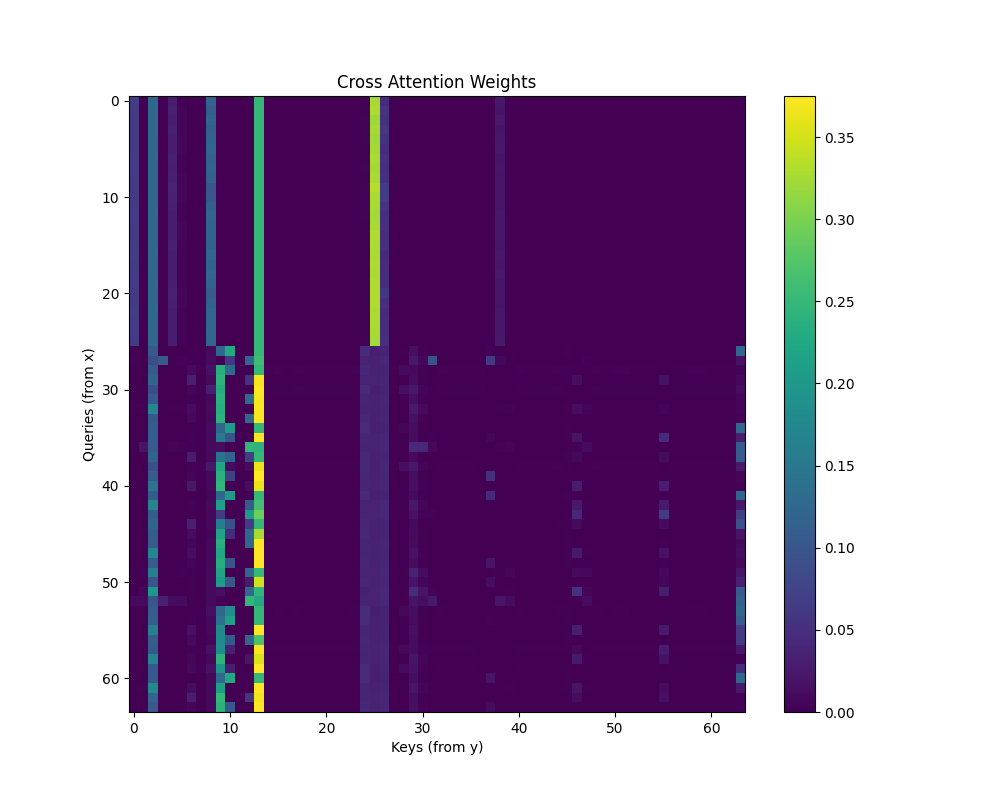}
    \end{subfigure}

    \vspace{1em}

    \begin{subfigure}[t]{0.6\textwidth} 
        \centering
        \includegraphics[width=\textwidth]{vitmae_plot_0035.jpg}
    \end{subfigure}
    \hfill
    \begin{subfigure}[t]{0.38\textwidth} 
        \centering
        \includegraphics[width=\textwidth, height=0.6\textwidth, keepaspectratio]{vitmae_attention_0035.png}
    \end{subfigure}

    \vspace{1em}

    \begin{subfigure}[t]{0.6\textwidth} 
        \centering
        \includegraphics[width=\textwidth]{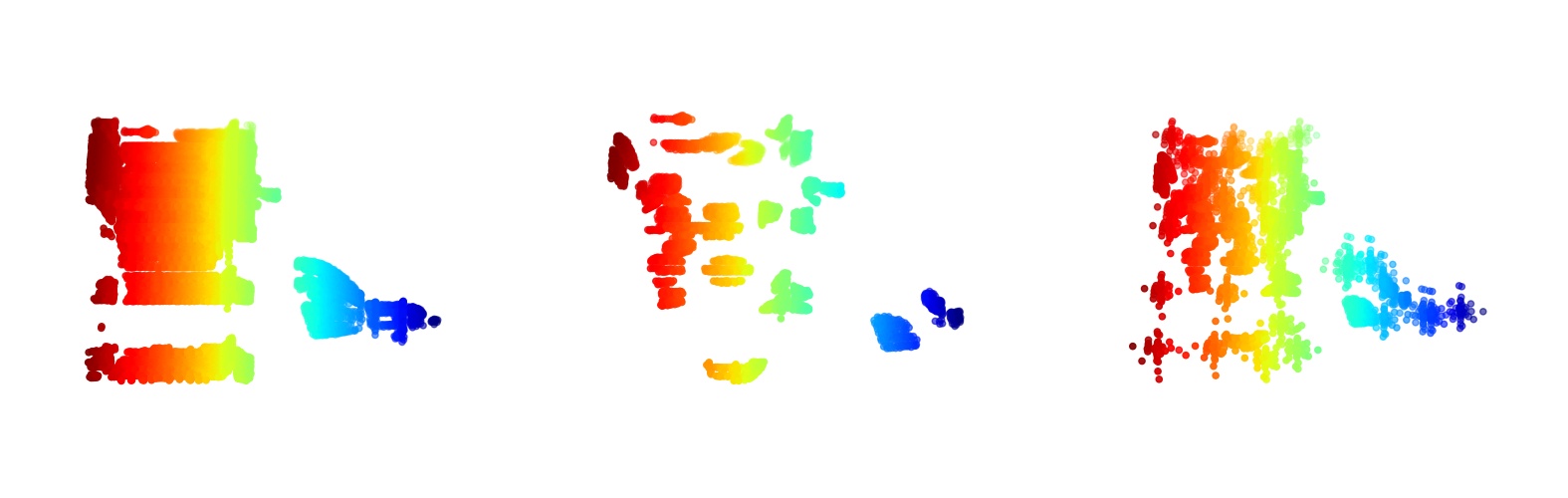}
    \end{subfigure}
    \hfill
    \begin{subfigure}[t]{0.38\textwidth} 
        \centering
        \includegraphics[width=\textwidth, height=0.6\textwidth, keepaspectratio]{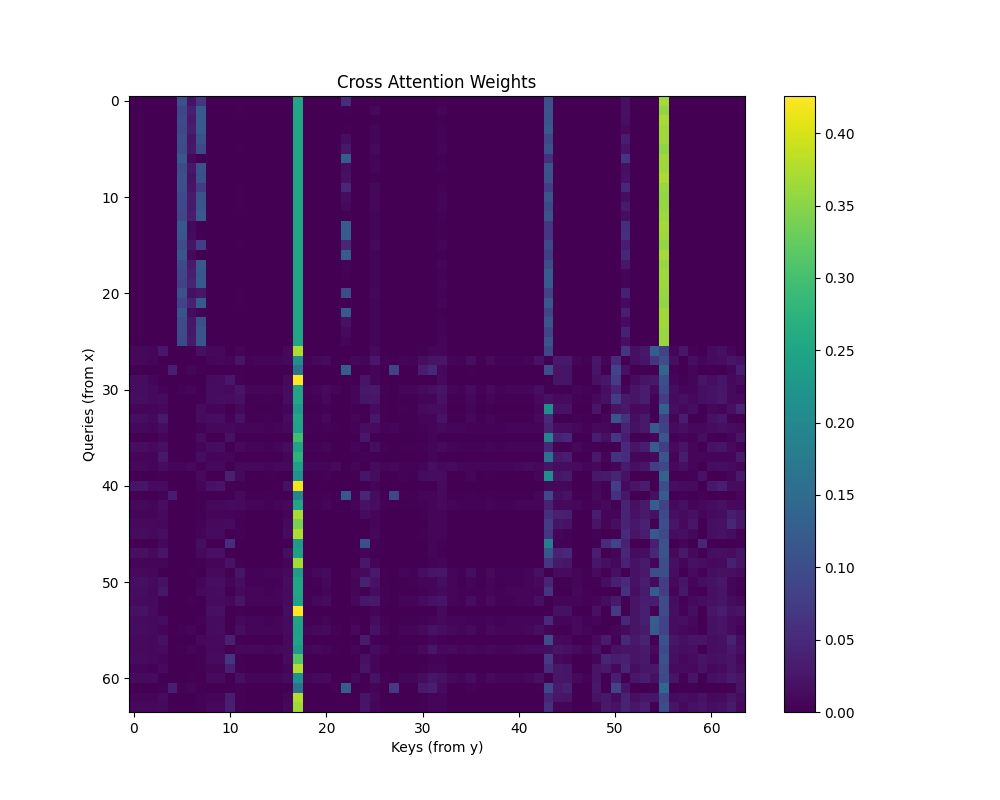}
    \end{subfigure}

    \vspace{1em}

    \begin{subfigure}[t]{0.6\textwidth} 
        \centering
        \includegraphics[width=\textwidth]{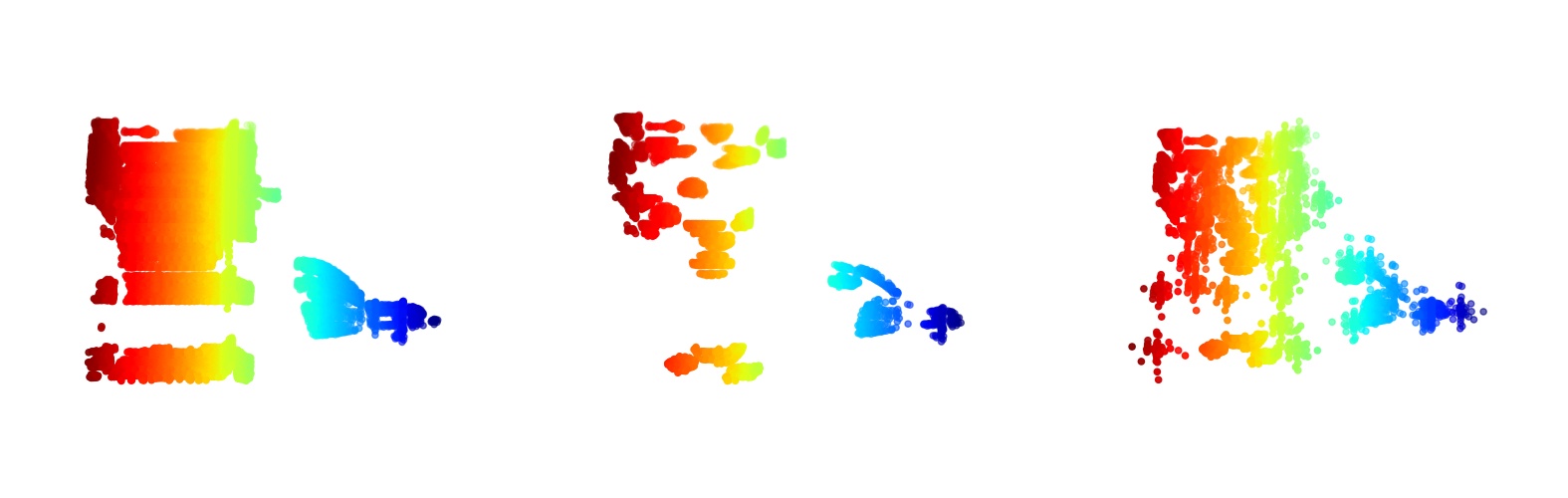}
    \end{subfigure}
    \hfill
    \begin{subfigure}[t]{0.38\textwidth} 
        \centering
        \includegraphics[width=\textwidth, height=0.6\textwidth, keepaspectratio]{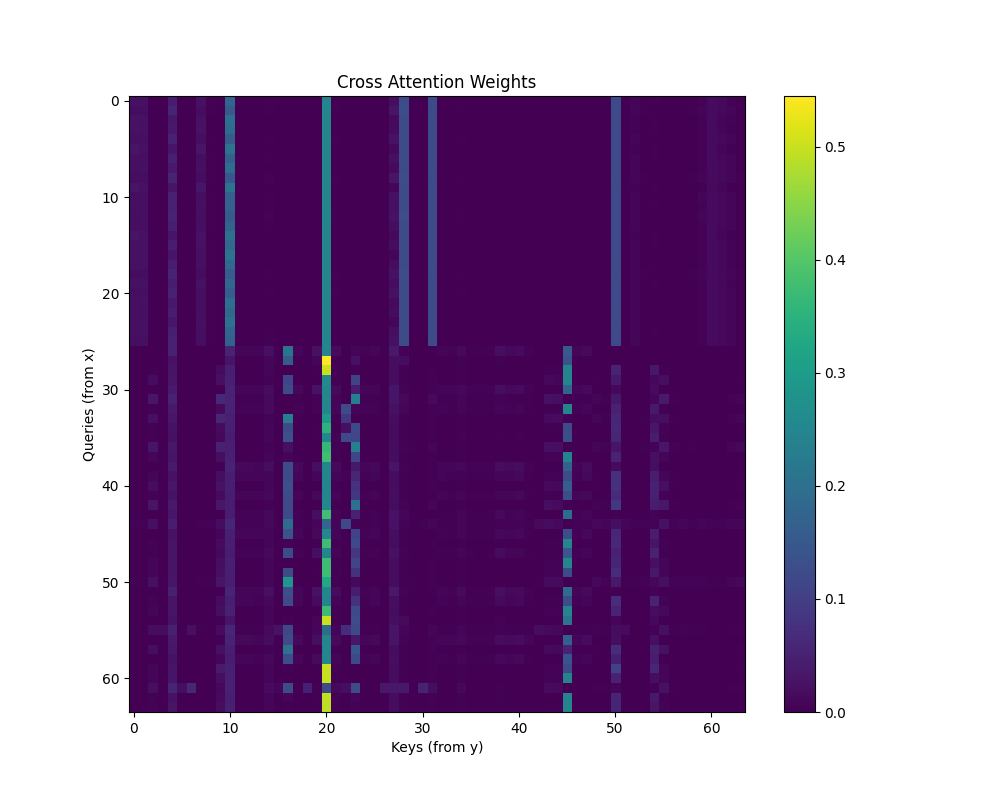}
    \end{subfigure}

    \vspace{1em}

    \begin{subfigure}[t]{0.6\textwidth} 
        \centering
        \includegraphics[width=\textwidth]{vitmae_plot_0088.jpg}
    \end{subfigure}
    \hfill
    \begin{subfigure}[t]{0.38\textwidth} 
        \centering
        \includegraphics[width=\textwidth, height=0.6\textwidth, keepaspectratio]{vitmae_attention_0088.png}
    \end{subfigure}

    \caption{Further ObitoNet/ViTMAE model sample outputs.}
    \label{fig:obitonet_vitmae_outputs}
\end{figure}

\subsection{ObitoNet/Large}
\begin{figure}[H]
    \centering

    \begin{subfigure}[t]{0.6\textwidth} 
        \centering
        \includegraphics[width=\textwidth]{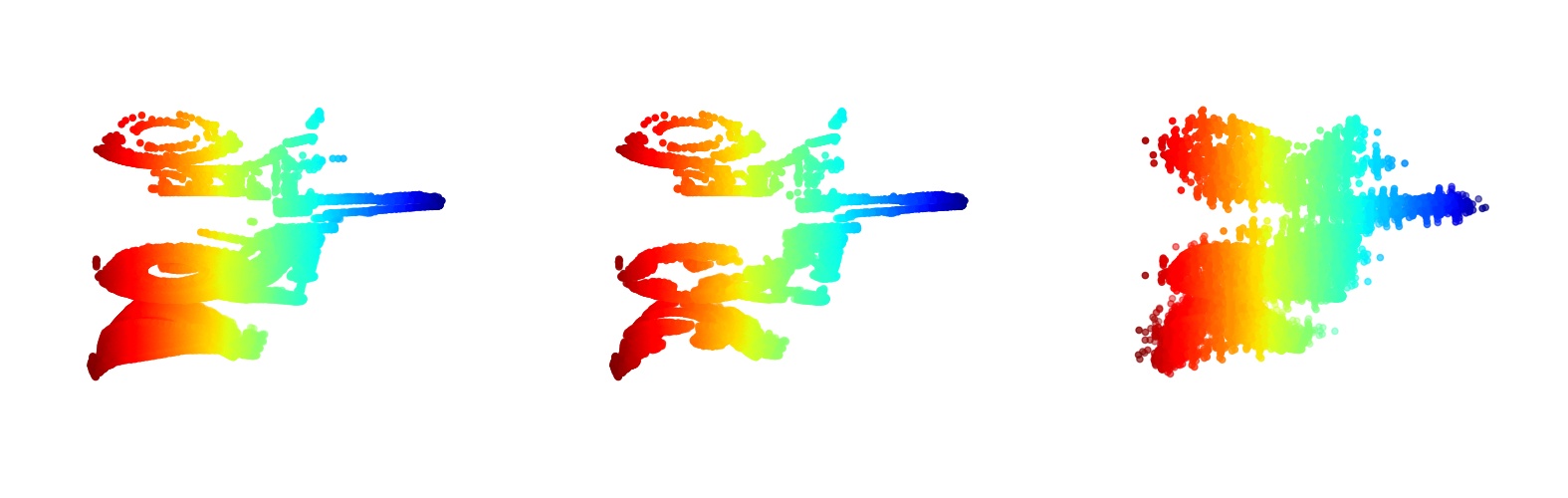}
    \end{subfigure}
    \hfill
    \begin{subfigure}[t]{0.38\textwidth} 
        \centering
        \includegraphics[width=\textwidth, height=0.6\textwidth, keepaspectratio]{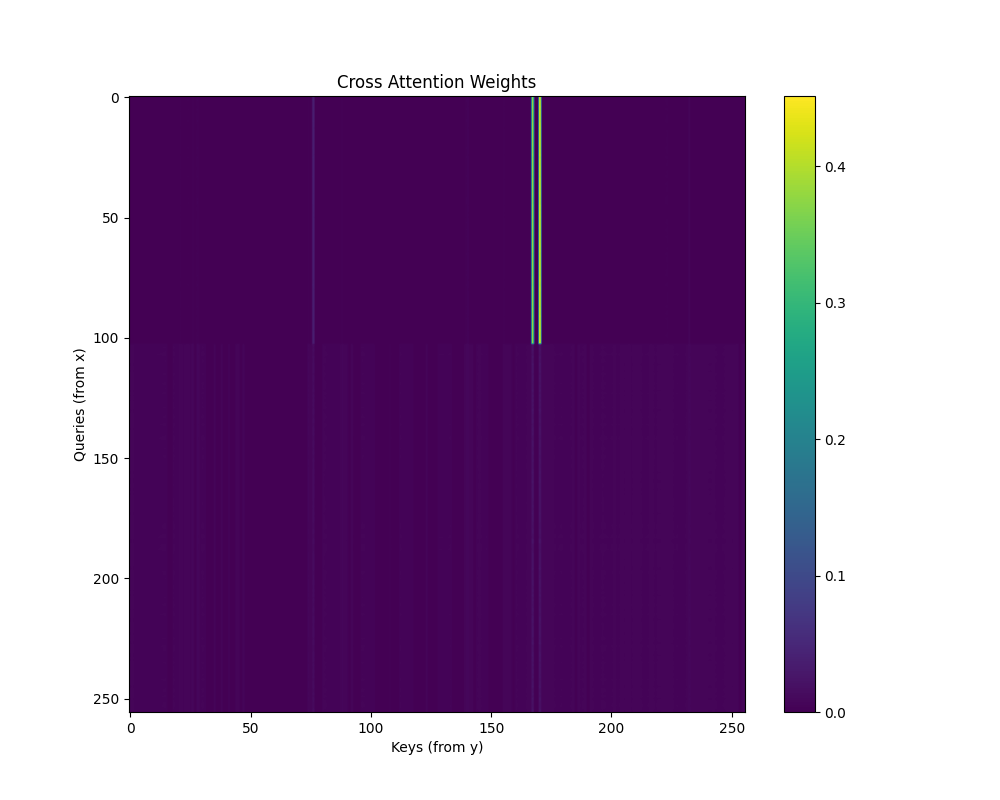}
    \end{subfigure}

    \vspace{1em}

    \begin{subfigure}[t]{0.6\textwidth} 
        \centering
        \includegraphics[width=\textwidth]{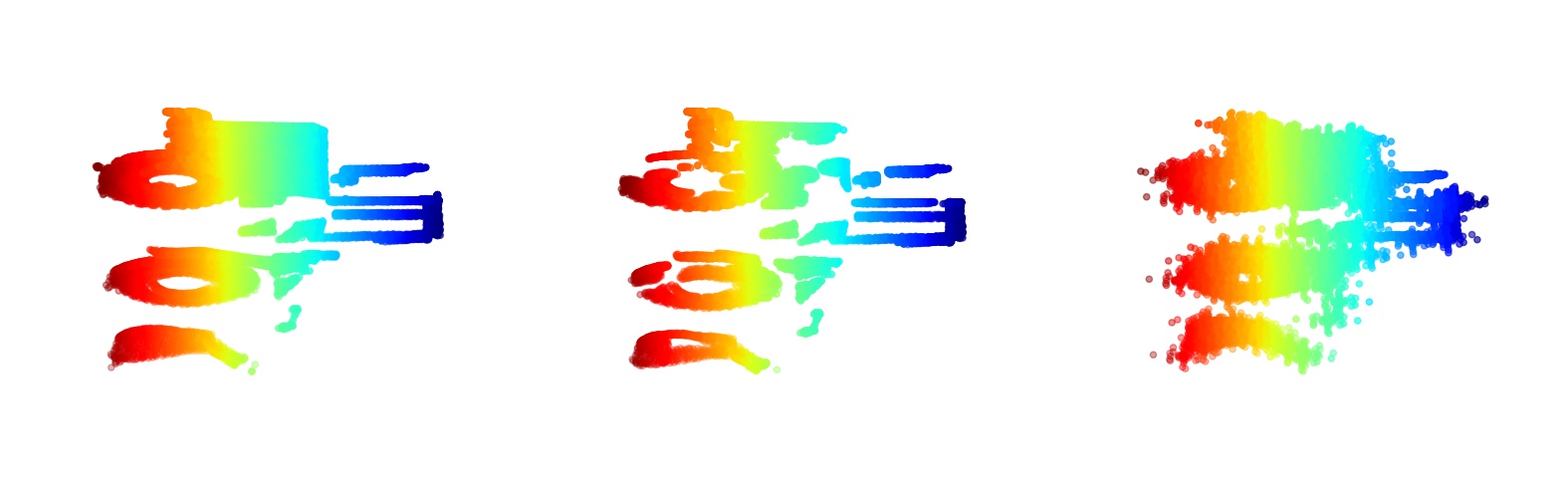}
    \end{subfigure}
    \hfill
    \begin{subfigure}[t]{0.38\textwidth} 
        \centering
        \includegraphics[width=\textwidth, height=0.6\textwidth, keepaspectratio]{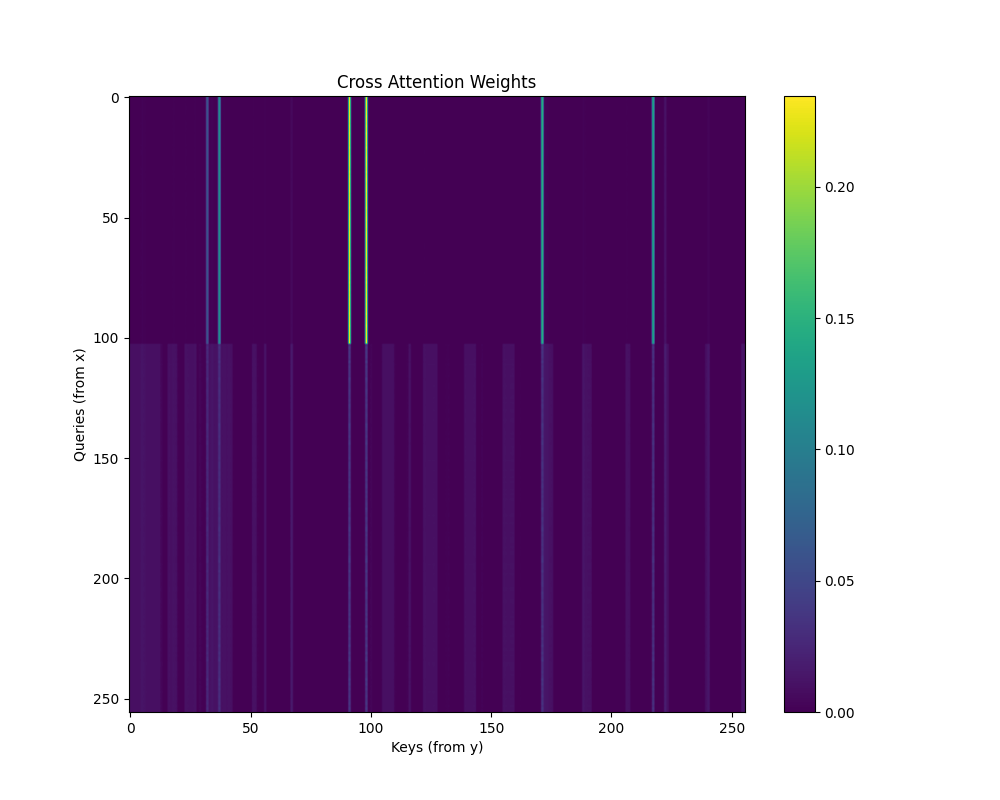}
    \end{subfigure}

    \vspace{1em}

    \begin{subfigure}[t]{0.6\textwidth} 
        \centering
        \includegraphics[width=\textwidth]{large_plot_0035.jpg}
    \end{subfigure}
    \hfill
    \begin{subfigure}[t]{0.38\textwidth} 
        \centering
        \includegraphics[width=\textwidth, height=0.6\textwidth, keepaspectratio]{large_attention_0035.png}
    \end{subfigure}

    \vspace{1em}

    \begin{subfigure}[t]{0.6\textwidth} 
        \centering
        \includegraphics[width=\textwidth]{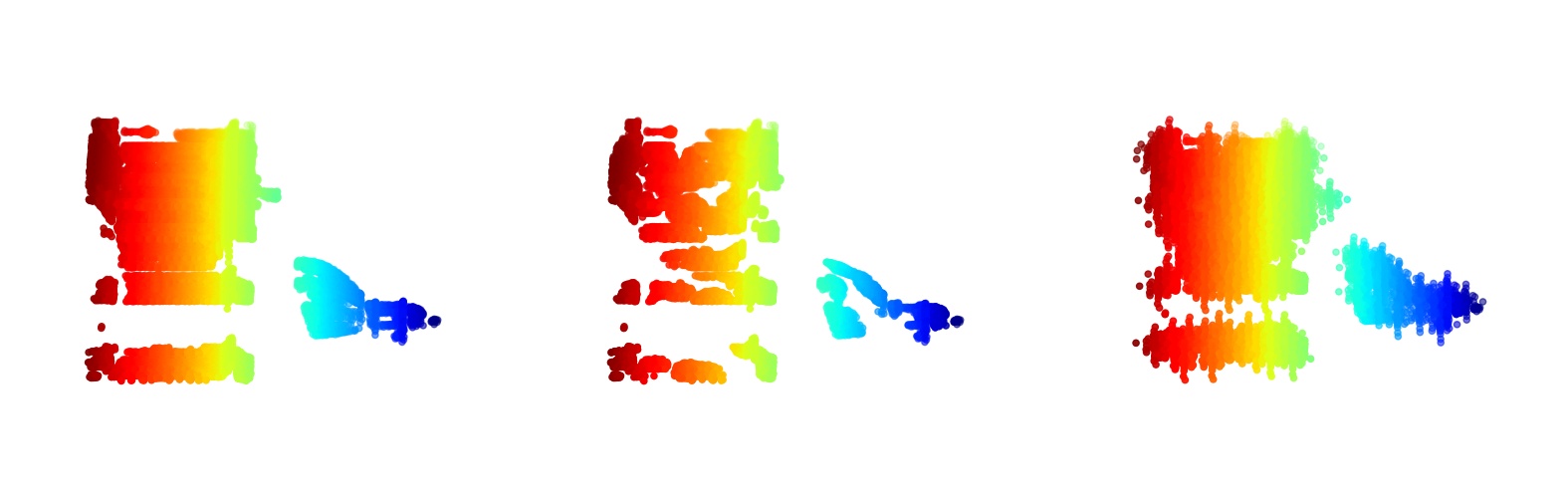}
    \end{subfigure}
    \hfill
    \begin{subfigure}[t]{0.38\textwidth} 
        \centering
        \includegraphics[width=\textwidth, height=0.6\textwidth, keepaspectratio]{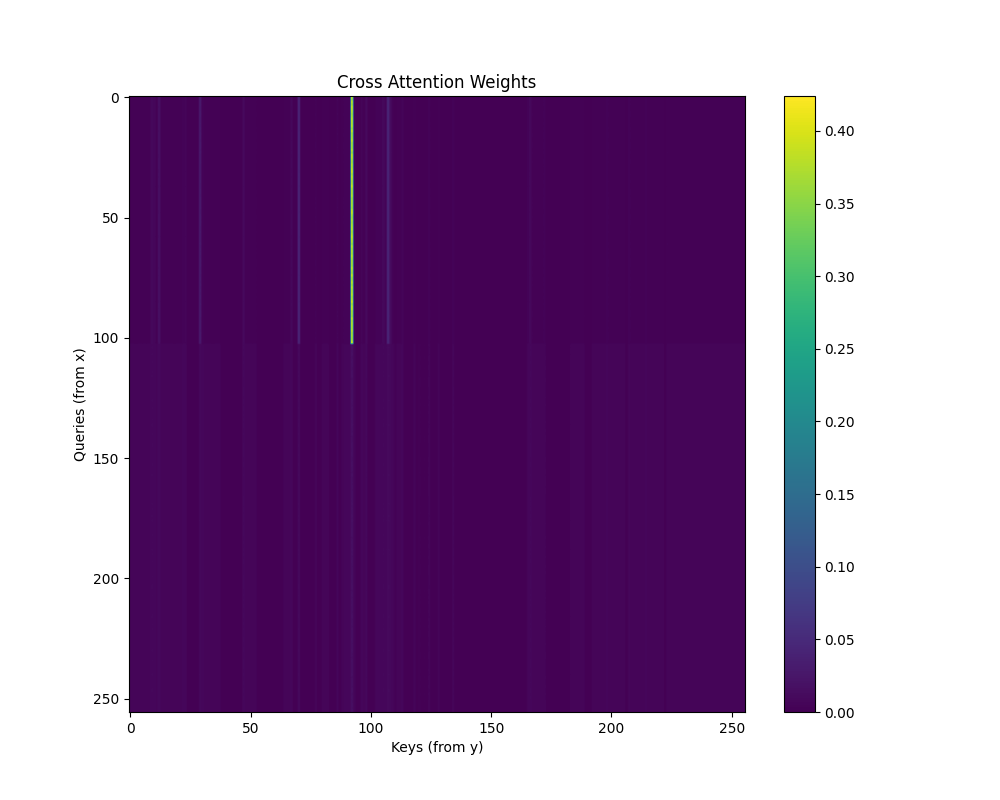}
    \end{subfigure}

    \vspace{1em}

    \begin{subfigure}[t]{0.6\textwidth} 
        \centering
        \includegraphics[width=\textwidth]{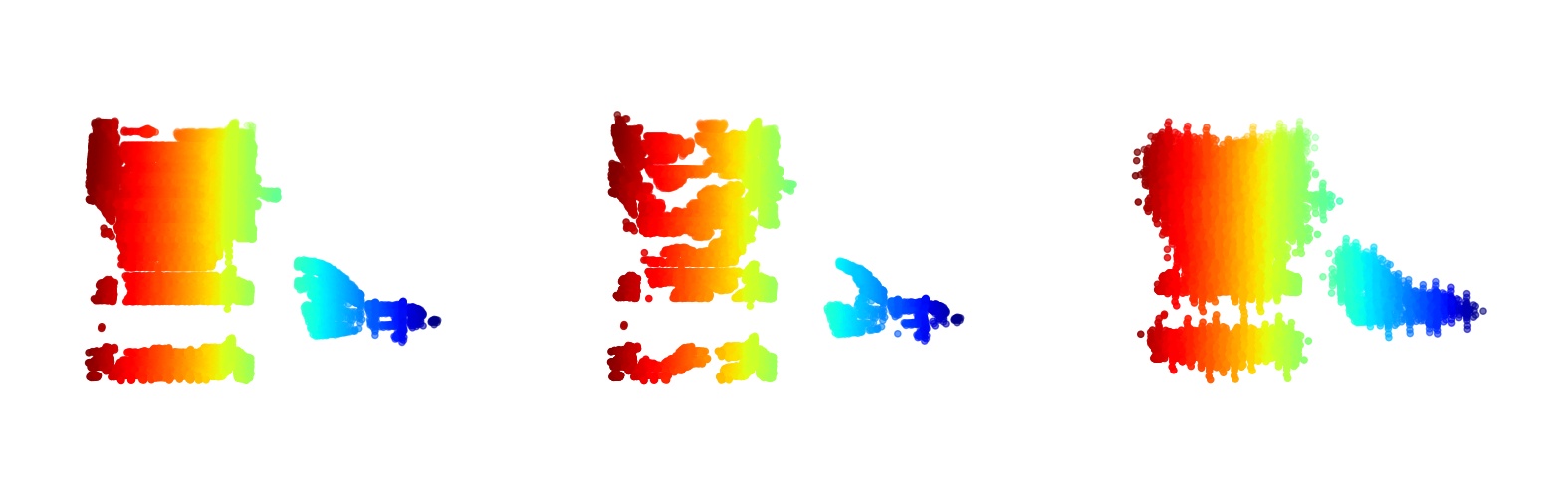}
    \end{subfigure}
    \hfill
    \begin{subfigure}[t]{0.38\textwidth} 
        \centering
        \includegraphics[width=\textwidth, height=0.6\textwidth, keepaspectratio]{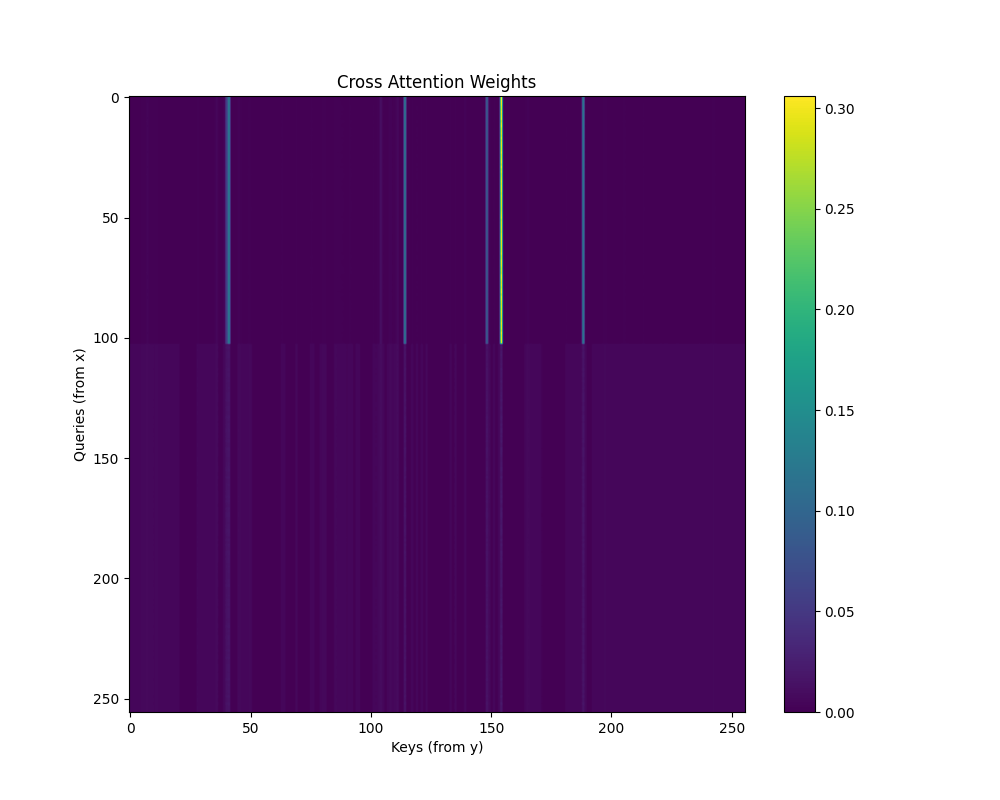}
    \end{subfigure}

    \vspace{1em}

    \begin{subfigure}[t]{0.6\textwidth} 
        \centering
        \includegraphics[width=\textwidth]{large_plot_0088.jpg}
    \end{subfigure}
    \hfill
    \begin{subfigure}[t]{0.38\textwidth} 
        \centering
        \includegraphics[width=\textwidth, height=0.6\textwidth, keepaspectratio]{large_attention_0088.png}
    \end{subfigure}

    \caption{Further ObitoNet/Large model sample outputs.}
    \label{fig:obitonet_large_outputs}
\end{figure}

\end{document}